\documentclass{article}
\usepackage{adjustbox}
\usepackage{microtype}
\usepackage{graphicx}
\usepackage{subfigure}
\usepackage{xcolor,colortbl}
\usepackage{booktabs}
\definecolor{LightCyan}{rgb}{0.88,1,1}
\newcolumntype{a}{>{\columncolor{LightCyan}}c}
\usepackage{hyperref}
\usepackage{makecell}
\usepackage{longtable}
\usepackage{multirow}
\usepackage{colortbl}

\usepackage{arydshln}
\usepackage[accepted]{icml2024}

\definecolor{ceruleanblue}{rgb}{0.16, 0.32, 0.75}
\definecolor{lightorange}{RGB}{255,223,155}
\usepackage{amsmath}
\usepackage{amssymb}
\usepackage{mathtools}
\usepackage{amsthm}
\usepackage{seqsplit}
\usepackage{inconsolata}
\DeclareMathOperator*{\argmax}{arg\,max}
\DeclareMathOperator*{\argmin}{arg\,min}

\usepackage[capitalize,noabbrev]{cleveref}
\theoremstyle{plain}

\theoremstyle{definition}

\theoremstyle{remark}

\usepackage{xcolor}

\definecolor{ao(english)}{rgb}{0.0, 0.5, 0.0}

\icmltitlerunning{Graph-enhanced Large Language Models in Asynchronous Plan Reasoning}

\begin{document}

\twocolumn[
\icmltitle{Graph-enhanced Large Language Models in Asynchronous Plan Reasoning}

\begin{icmlauthorlist}
\icmlauthor{Fangru Lin}{ox}
\icmlauthor{Emanuele La Malfa}{ox,at}
\icmlauthor{Valentin Hofmann}{ox,ai,lmu}
\icmlauthor{Elle Michelle Yang}{ox}\\
\icmlauthor{Anthony G. Cohn}{at,leeds}
\icmlauthor{Janet B. Pierrehumbert}{ox}
\end{icmlauthorlist}

\icmlaffiliation{ox}{University of Oxford}
\icmlaffiliation{ai}{Allen Institute for AI}
\icmlaffiliation{at}{Alan Turing Institute}
\icmlaffiliation{lmu}{LMU Munich}
\icmlaffiliation{leeds}{University of Leeds}

\icmlcorrespondingauthor{Fangru Lin}{fangru.lin@ling-phil.ox.ac.uk}

\icmlkeywords{Graph, Large Language Model, Planning}

\vskip 0.3in
]

\printAffiliationsAndNotice{} 

\begin{abstract}
Planning is a fundamental property of human intelligence. Reasoning about asynchronous plans is challenging since it requires sequential and parallel planning to optimize time costs. Can large language models (LLMs) succeed at this task? Here, we present the first large-scale study investigating this question. We find that a representative set of closed and open-source LLMs, including GPT-4 and LLaMA-2, behave poorly when not supplied with illustrations about the task-solving process in our benchmark AsyncHow. We propose a novel technique called \emph{Plan Like a Graph} (PLaG) that combines graphs with natural language prompts and achieves state-of-the-art results. We show that although PLaG can boost model performance, LLMs still suffer from drastic degradation when task complexity increases, highlighting the limits of utilizing LLMs for simulating digital devices. We see our study as an exciting step towards using LLMs as efficient autonomous agents. Our code and data are available at \url{https://github.com/fangru-lin/graph-llm-asynchow-plan}.
\end{abstract}

\section{Introduction}
\label{sec:intro}
\begin{figure}[!ht]
    \centering
    \includegraphics[width=0.48\textwidth]{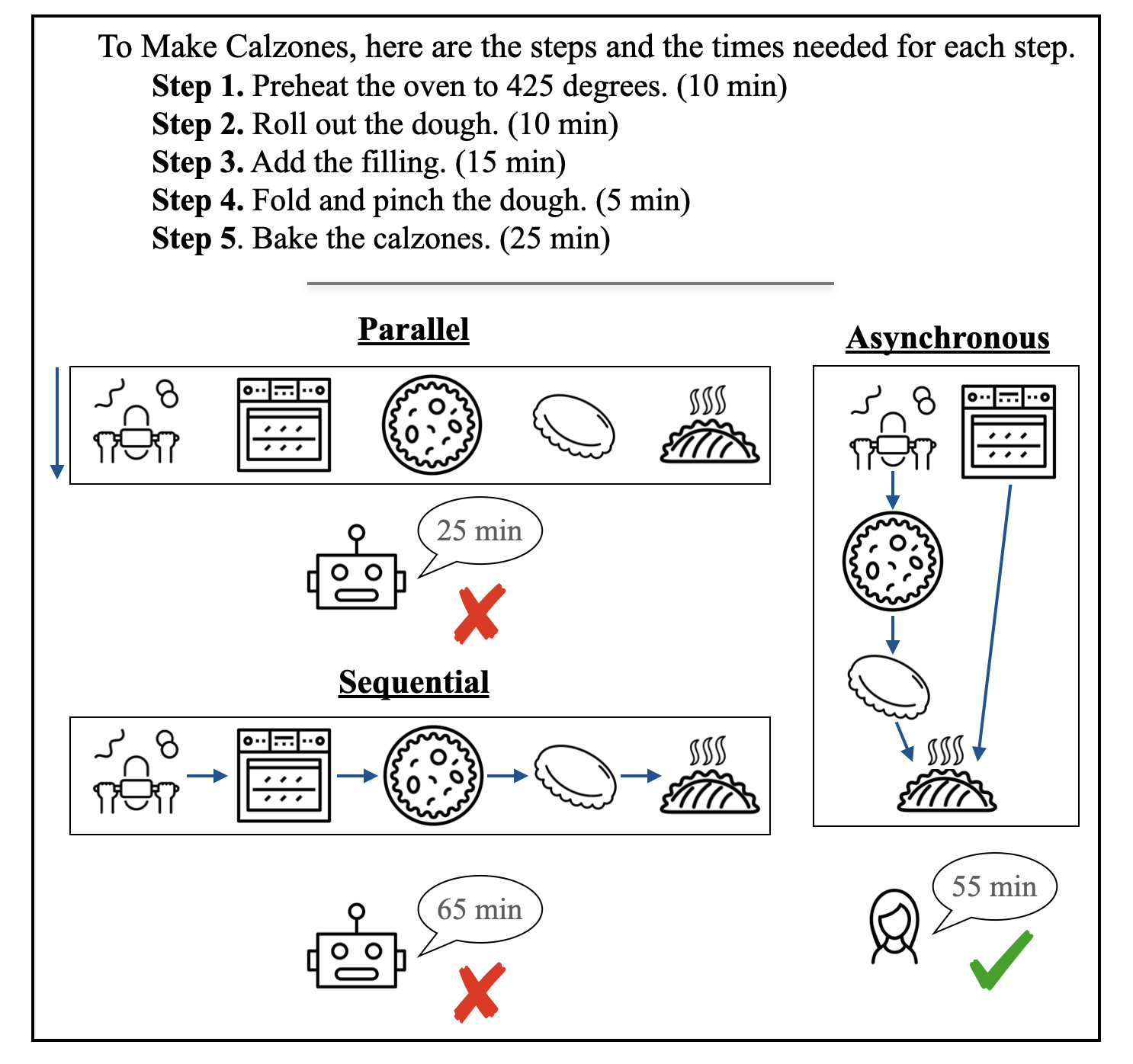}
    \vspace{-5mm}
    \caption{A planning task (top) can be executed sequentially, in parallel, or asynchronously. \textcolor{ceruleanblue}{Blue} arrows denote action ordering constraints. Although complete parallelism is logically the most time-efficient strategy, it results in invalid reasoning steps (e.g. \texttt{`Baking'} cannot happen at the same time with \texttt{`Rolling the dough'}); at the same time, sequentially executing each task negatively affects efficiency. Given infinite resources, an optimal (asynchronous) plan should parallelize actions wherever possible.}
    \label{fig:task_desc}
    \vspace{-5mm}
\end{figure}
As large language models (LLMs) show unprecedented capabilities, claims surge that artificial general intelligence is close \cite{bubeck2023sparks}. Planning is an important property of human intelligence \cite{sternberg1984toward, colom2010human}, and it is also vital in many downstream tasks such as developing autonomous robotic agents \cite{huang2022language,shinn2023reflexion}. While symbolic processors have been historically used for handling plan design~\cite{fikes1971strips, mcdermott20001998}, LLMs have recently emerged as a relevant complementary approach~\cite{ahn2022can, dagan2023dynamic, song2023llm}. Although LLMs generate reasonable elementary planning steps when informed with appropriate guidance~\cite{pmlr-v162-huang22a, yuan-etal-2023-distilling}, they cannot combine those units effectively and develop optimal plans without external processors \cite{silver2022pddl, dagan2023dynamic, yang-etal-2023-coupling}. This might be an issue if LLMs are deployed for related tasks.
\begin{figure*}[!ht]
    \centering
    \includegraphics[scale=0.25]{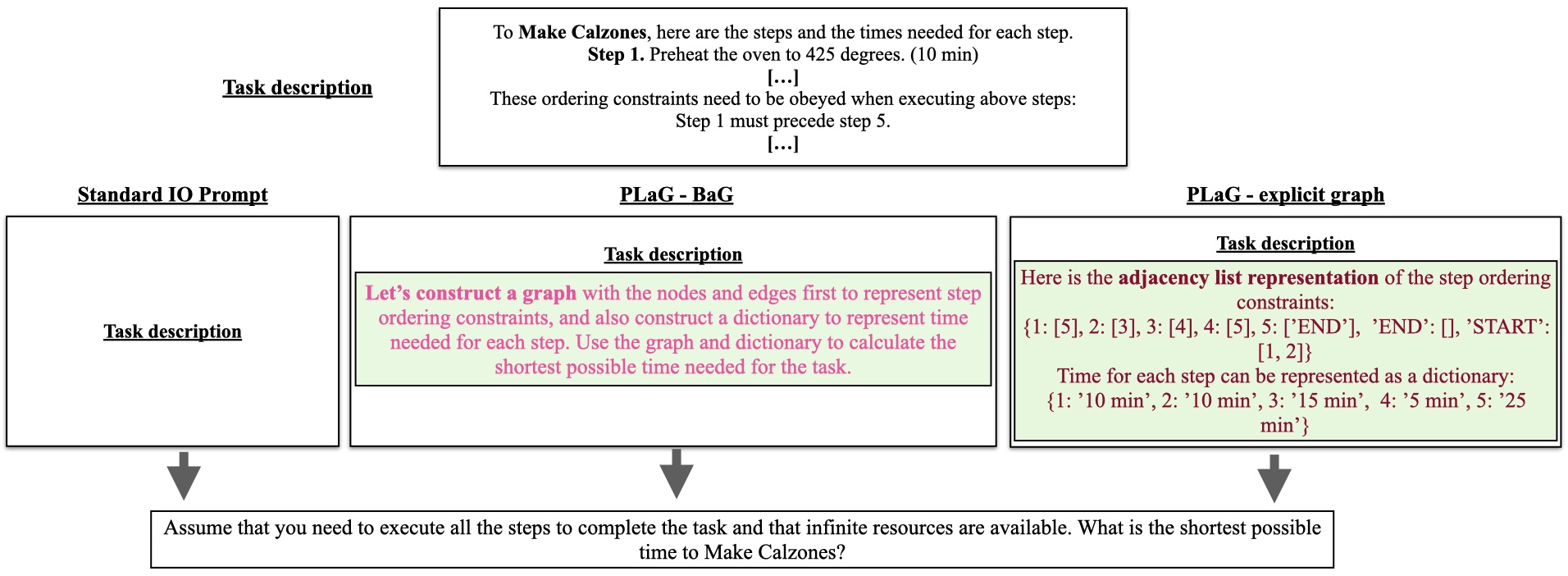}
    \caption{Comparing standard Input-Output (IO) prompting with our method (PLaG). Here, we illustrate PLaG (explicit graph) with an adjacency list, but it can be of any graph type in practice. The standard IO method is similarly deployed in zero-shot, zero-shot + CoT, \emph{k}-shot, \emph{k}-shot + CoT in this paper. Please refer to Appendix \ref{sec:prompt-bench} for more details.}
    \label{fig:plag-illustration}
\end{figure*}

\begin{figure}[!t]
    \centering
    \includegraphics[width=0.9\linewidth]{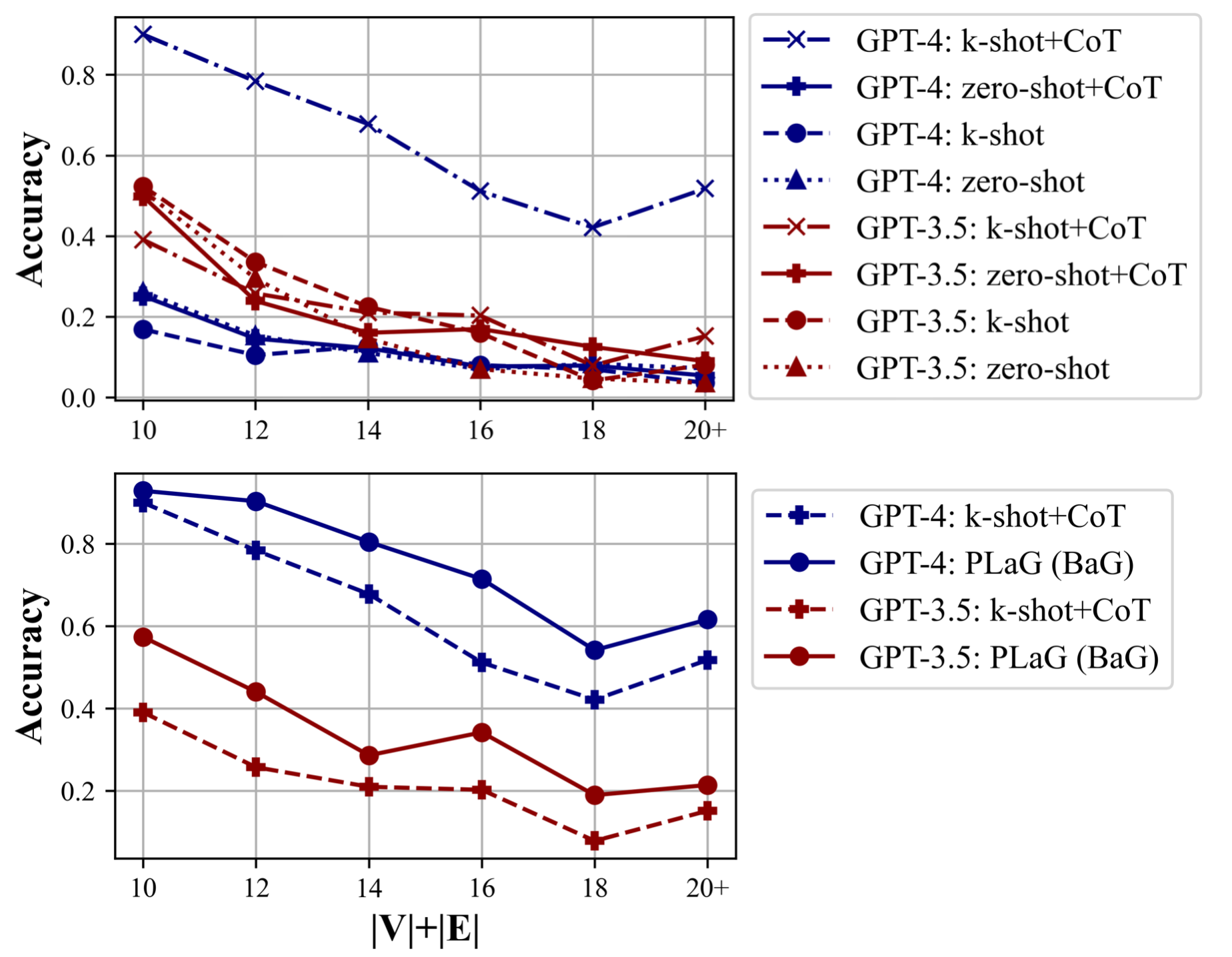}
    \vspace{-5mm}
    \caption{\small GPT-3.5 and GPT-4 accuracy as a function of asynchronous planning task complexity $|V|+|E|$ (see Section \ref{sec:formalism}), after binning results by width of 2. The upper figure plots the performance of methods without PLaG (our method), and the lower plot displays the best method with/without PLaG.}
    \label{fig:acc-ve}
\end{figure}

This work explores the reasoning ability of LLMs in naturalistic asynchronous planning, which we define as complex planning tasks involving both sequential and parallel actions. Given a set of steps for a task, the time required for each step, and step ordering constraints, we ask whether LLMs can compute the shortest possible time needed for an optimal plan for the task (Figure~\ref{fig:task_desc}). We note that asynchronous planning problems involve (i) time summation (correctly adding time durations), (ii) time comparison (correctly making time duration comparisons), and (iii) constrained reasoning (correctly solving constrained optimization problems) --- this compositionality of skills makes asynchronous planning a challenging task, and it is yet unclear whether LLMs are capable of solving it.
To enable a large-scale evaluation of LLMs, we automatically generate a new benchmark, \textbf{Async}hronous Wiki\textbf{How} (AsyncHow), with 1.6K high-quality instances for real-life tasks.

We use AsyncHow to evaluate GPT-3.5-turbo (GPT-3.5), GPT-4~\cite{OpenAI2023GPT4TR}, Cohere Command\footnote{https://cohere.com/models/command}, LLaMA-2-70B-chat \cite{touvron2023llama}, and Mistral-7B-Instruct \citep[v0.2;][]{jiang2023mistral} on asynchronous planning. We find that while GPT-4 with few-shot task solution illustrations dominates other models in terms of accuracy, all models perform poorly without illustrations about how to solve the task. However, even with few-shot illustrations, model performance is unsatisfactory, with the LLMs failing on instances that are trivial for humans. To remedy this, we propose a novel prompting technique, namely \textbf{P}lan \textbf{L}ike \textbf{a} \textbf{G}raph (PLaG; Figure~\ref{fig:plag-illustration}), to instruct models to represent a planning problem like a graph. By converting naturalistic questions to equivalent graph problems, we find that our method boosts the performance of all tested models. Moreover, it can be applied \textbf{off the shelf} to models such as GPT-4 to achieve new state-of-the-art (SOTA) results and consistently improve on all task complexity levels (Figure~\ref{fig:acc-ve}, lower). Nonetheless, we find that the improved models still suffer from drastic performance degradation on complex planning tasks.

In summary, the main contributions of this paper are:
\begin{itemize}
    \item We automatically generate a high-quality naturalistic benchmark for asynchronous plan reasoning, \mbox{AsyncHow}, and open-source it. 
    \item We show that LLMs cannot efficiently execute asynchronous plans unless they are supplied with detailed solution illustrations.
    \item We provide a formalism to define the complexity of naturalistic asynchronous planning tasks, which successfully predicts LLMs' performance trends.
    \item  We propose PLaG, an off-the-shelf method to consistently boost SOTA model performance across all considered task complexities.
    \item We show that despite the performance boost, SOTA LLMs suffer from drastic degradation with increasing task complexity, which indicates that there are limits to using LLMs as digital devices.
\end{itemize}

The paper is structured as follows. Section~\ref{sec:formalism} introduces our asynchronous planning task, and formally defines its complexity as an optimization problem as well as our technique. Section~\ref{sec:dataset} describes how we generate the benchmark. Section~\ref{sec:experiment} lays out the experimental setting and overall results. The results are then analyzed in more detail in Section \ref{sec:analysis}. We review relevant related works and conclude the article with our main contributions in Sections~\ref{sec:rw} to~\ref{sec:impact}. 

\section{Preliminaries: Naturalistic Asynchronous Planning}
\label{sec:formalism}

\begin{figure}
    \centering
    \includegraphics[width=0.9\linewidth]{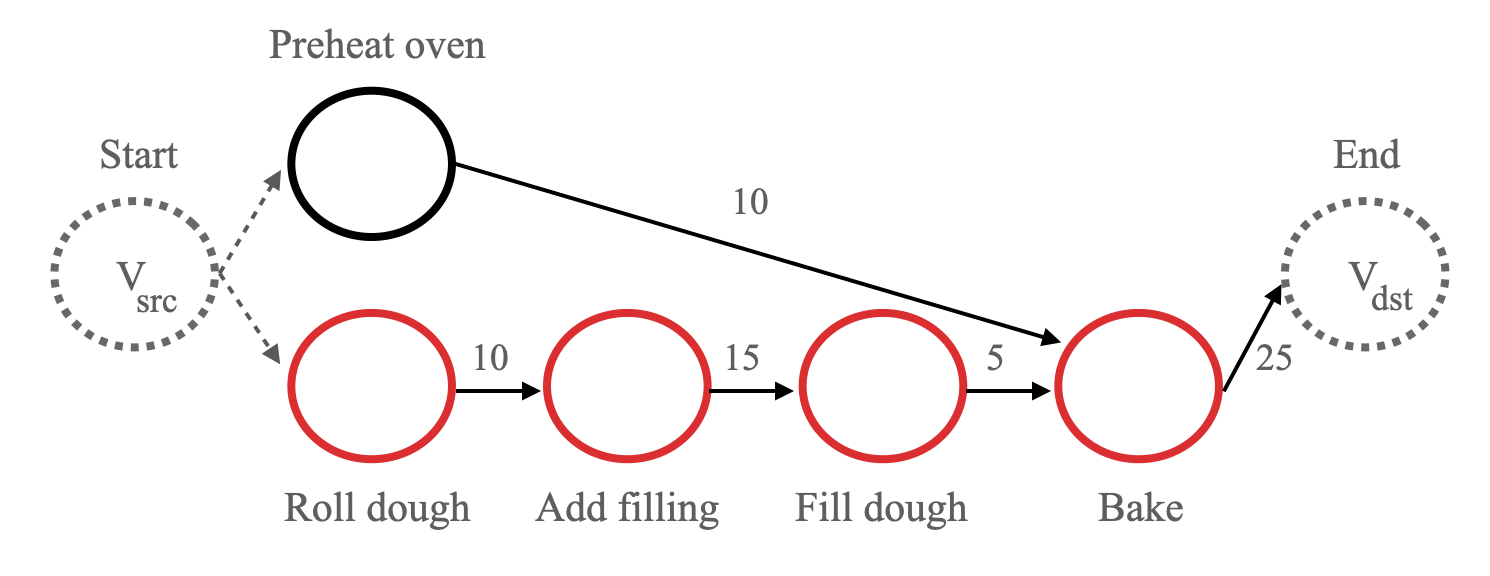}
    \caption{The series-parallel DAG used to solve the planning task in Figure~\ref{fig:task_desc}. The path for calculating optimal time duration is highlighted in \textcolor{red}{red}.}
    \label{fig:dag_task_desc}
\end{figure}

We define our task as follows: assuming infinite resources (e.g., as many agents and tools as needed to achieve optimal parallelism are available) for a naturalistic task with a set of compulsory steps, the time needed for each step, and step ordering constraints, we assess whether LLMs can compute the optimal time needed for the task. Formally, we can cast this as the problem to find the longest path on a Directed Acyclic Graph (DAG, Figure \ref{fig:dag_task_desc}). A key advantage of doing so is that we can easily estimate the complexity of our task despite the fact that it is a natural language processing problem. This distinguishes our work from many other studies. We empirically prove that our complexity measure predicts LLM behavior in all prompt settings (Figure~\ref{fig:acc-ve} and Section \ref{sec:acc-complexity}), with variance explained in Section \ref{sec:qual_study}.

Since our task is essentially similar to DAG search, it sheds light on the limits of LLMs as \emph{digital} devices~\cite{la2024code} and, specifically, as solvers of discrete optimization problems on graphs~\cite{wang2023can}. It also serves as (i) an example of an optimal succinct routine that an LLM might be able to implement internally~\cite{weiss2021thinking} to solve a planning problem and (ii) a baseline to measure the loss induced by specifying a problem in natural language, namely an LLM's \emph{language divide}. 

\subsection{Complexity of Naturalistic Planning}
We briefly introduce the complexity measure for our task in this subsection,  which we will later show correlates with LLM behavior. With infinite resources, the formalism of a DAG captures the complexity of finding the optimal execution order of compulsory actions $a$ in a plan $P$ to minimize the time cost $TC(P)$. A DAG $G(P)$ representing $P$ can be defined as~$G(P) = \langle V, E, w \rangle$, where $V$ is a set of nodes $v$, each representing an action $a$ in the planning problem, including auxiliary START ($v_{\text{src}}$) and END ($v_{\text{dst}}$). 
$E$ is a directed set of flow relations $e$ representing ordering constraints, while $w$ is a function that assigns a weight to all edges in the graph $w: E \rightarrow \mathbb{R}^+$. Each flow relation $e_{i,j}$ is associated with a positive number  $w(e_{i,j})$ to express that node/action $v_{i}$ is connected to node/action $v_{j}$ and requires $w(e_{i,j})$ time to be completed. The edges also represent causal links in that the precondition for an action/node $a$ is met if and only if all actions/nodes linked to and preceding $a$ are performed. For simplicity, we denote $G(P)$ as $G$ in the remaining part of the paper.

In this setting, finding the time cost for an optimal plan  $P^*$ in a planning problem is equivalent to finding the longest path $G^*$ on $G$ and can be cast as the following optimization problem on a subgraph $G'= \langle V', E', w\rangle, \ G' \subseteq G$. Exhaustively searching a graph and comparing every path's length can deterministically find the gold answer. On series-parallel graphs~\cite{eppstein1992parallel}, which are sufficient to describe planning tasks with infinite resources, the average time complexity is $\mathcal{O}(|V|+|E|)$~\cite{takamizawa1982linear}, i.e., it is linear with respect to the number of nodes and edges in $G$.\footnote{While we assume infinite resources to complete a planning task, the natural extension to the case of finite resources (i.e., not all independent actions can be parallelized) is better captured by the formalism of a Petri net (or discrete-time Markov chains with constraints). We introduce the current formalism and Petri net in more detail in Appendix~\ref{sec:petri_net}.} We define our task complexity accordingly.

\begin{figure*}[t!]
    \centering
    \includegraphics[scale=0.53]{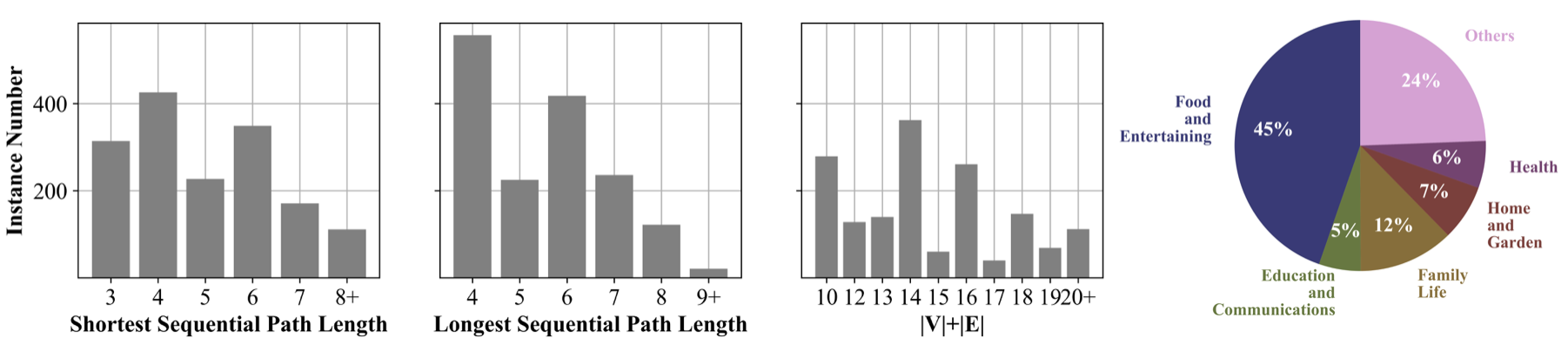}
    \caption{Overview of the AsyncHow benchmark. The three bar charts on the left display the instance numbers for the shortest/longest sequential path length and $|V|+|E|$ in different plans. The pie chart on the right shows the topic distribution in our dataset. See Appendix~\ref{sec:topic-assignment} for details about the topic assignment.}
    \label{fig:data-distribution}
\end{figure*}

\subsection{Method: Plan Like a Graph}
In our work, we propose a novel prompting technique \textbf{P}lan \textbf{L}ike \textbf{a} \textbf{G}raph (PLaG, Figure \ref{fig:plag-illustration}).  Taking inspiration from Discourse Representation Theory~\cite{wolf2004discourse} and relevant works on graphical prompt representations~\cite{fatemi2023talk, wang2023can}, PLaG includes a graph representation in the prompt, where we give models \emph{k}-shot illustrations with graphs describing the task and instruct them to either reason based on a given graph (i.e., explicit graph) or to generate a graph themselves and then reason about it \citep[i.e., \textbf{B}uild \textbf{a} \textbf{G}raph/BaG; ][]{wang2023can}. We instruct models to produce graph representations of the naturalistic question and then use the information to solve relevant tasks.

\section{The AsyncHow Benchmark for Planning}
\label{sec:dataset}

\begin{table*}[t!]
    \caption{Comparison of our step dependency annotation for the ProScript dev and test set, with mean and standard deviation performance on three randomized experiments (100 instances per experiment).}
    \centering
    \begin{tabular}{l|ccc|ccc}\hline
        &\multicolumn{3}{c}{dev (in-domain)}&\multicolumn{3}{|c}{test (cross-domain)}\\\hline
         &F1&P&R&F1&P&R\\\hdashline
         Humans&89.32&89.60&89.21&89.28&89.91&88.86\\
         GPT-4&89.80$_{\pm1.70}$&90.65$_{\pm1.36}$&89.30$_{\pm2.09}$&85.59$_{\pm2.92}$&85.95$_{\pm2.43}$&85.77$_{\pm3.56}$\\\hline
    \end{tabular}
    \label{tab:dep_val}
\end{table*}

Since there is no existing dataset appropriate for our defined task, we generate a new naturalistic asynchronous planning benchmark called \textbf{Async}hronous Wiki\textbf{How} (AsyncHow). This section describes and validates an \emph{automatic} method for generating this benchmark. With LLMs that consume new benchmarks at an unprecedented pace, our contribution goes beyond AsyncHow and can be used by practitioners to synthesize new datasets.

In addition to the existing data in ProScript \cite{sakaguchi-etal-2021-ProScript-partially}, an end-to-end human-annotated partial-order plan dataset\footnote{ProScript is similar to our dataset, but it is not suitable enough for our task. See Section~\ref{sec:rw} for discussions.}, we use WikiHow~\cite{koupaee2018wikihow,zhang-etal-2020-reasoning} to collect the planning tasks we need. In line with recent works, we use LLMs as data annotators \cite{gilardi2023chatgpt,huang2023chatgpt}. Specifically, we use the GPT models for part of pre-processing, time annotation, and step dependency annotation, as they exhibit impressive annotation capabilities~\cite{he2023annollm}. However, we would like to stress that (i) any LLM (or equivalent algorithmic procedure) can be used as an annotator and (ii) the LLM used to annotate is not involved in the ground truth answer generation, where we use deterministic procedures such as the longest path on a DAG. This means that the GPT models used for annotation should not be considered oracles in the benchmarking experiments. 

This process culminates in \textbf{AsyncHow}, a curated list of 1.6K data points for asynchronous planning. We provide an overview of the benchmark structure in Figure~\ref{fig:data-distribution}. We evaluate the dependency annotation quality automatically and the general generation quality with human annotators (Section~\ref{sec:quality-check}). We do not verify the time annotation because the task time estimation in a less grounded setting such as ours tends not to have a unique gold answer (e.g., \texttt{`finding a gym'} may take five minutes or a week to different people), and GPT-3.5 (the model we use for time annotation) is reported to be a reliable annotator for this task~\cite{jain2023language}. Furthermore, our interest is in assessing whether an LLM outputs the optimal plan for a task, and we expect end users to supply different time durations when querying a model.

We now briefly describe the data generation process, with more details in Appendix~\ref{sec:data-gen-app}. 

First, we \textbf{preprocess} the dataset to collect high-quality plans rated by WikiHow users. Then, given our task definition (e.g., all steps need to be executed, etc.), we filter out plans with optional steps and others that do not fit into our research goal by both matching keywords and prompting GPT-3.5 to answer relevant questions (e.g., \texttt{Are all steps needed in this plan?}). Then, we use GPT-3.5 to \textbf{estimate the time duration} per step and exclude instances whose step durations cannot be quantified numerically.

Next, we use GPT-4 to \textbf{annotate step dependencies} with the dot language. After removing redundant dependencies (e.g., in an answer saying \texttt{`step 1 $\rightarrow$ step 2', `step 2 $\rightarrow$ step 3', `step 1 $\rightarrow$ step 3'}, we remove \texttt{`step 1 $\rightarrow$ step 3'}), we keep data points that have at least four consistent answers that form asynchronous plans and discard the others.

After the above steps, we combine all asynchronous instances with complete time annotation for all meaningful steps in ProScript with our generated asynchronous instances from WikiHow, after which we obtain a collection of 1.6K instances. We then \textbf{generate natural language prompts} based on the task information in dot language, as users tend to use natural language descriptions to specify such a task. We have 10 trivially different plausible templates with their succinct use cases in our dataset (e.g. \texttt{`step 1 ->  step 2, step 1 -> step 3'} may be expressed as \texttt{`Step 1 must precede step 2, step 1 must precede step 3'}, and succinctly as \texttt{`Step 1 must precede step 2 and 3'}) to allow for relevant paraphrase robustness studies \cite{elazar2021measuring}.

Last, we \textbf{generate equivalent DAGs} representing the workflow and \textbf{compute the optimal time duration} for a plan by calculating the time duration for the longest one. Each planning task is eventually accompanied by four types of graph representations: the adjacency and the edge list, the adjacency matrix, and the compressed sparse row (csr), which can be used to aid LLMs in structural reasoning and assess LLMs' robustness against different graph representations. We do not further vary natural language representations for graphs because relevant investigations can be found in \citet{fatemi2023talk}.

\begin{table*}[t!]\centering
    \caption{Model accuracy in different settings on the AsyncHow benchmark. Model performances without our method are in plain background, while those with our method are in blue background. We mark the best performance per model in \textbf{bold}. Following \citet{dror-etal-2018-hitchhikers}, we use McNemar’s tests~\cite{mcnemar1947note} to obtain \emph{p}-values and Holm-Bonferroni method~\cite{holm1979simple} to correct them for each evaluation to test the statistical significance of performance difference between experiment with and without our proposed method. We denote with $\dag$ when the performances with PLaG are significantly better ($p<0.05$) than the best result without.}
    \centering
    \begin{tabular}{lccccaa}\toprule
    & \multicolumn{4}{c}{Without PLaG} & \multicolumn{2}{a}{With PLaG}\\
    \cmidrule(lr){2-5}\cmidrule(lr){6-7}
        Model & zero-shot & zero-shot + CoT & \emph{k}-shot & \emph{k}-shot + CoT& PLaG (explicit graph) & PLaG (BaG) \\ \hline
        GPT-4 & 0.130 &0.129 &0.107 &0.657 &0.730$^\dag$ &\textbf{0.777}$^\dag$ \\
        GPT-3.5 & 0.199 &0.224 &0.248 &0.226 &0.290$^\dag$ &\textbf{0.355}$^\dag$ \\
        Command & 0.078 & 0.015& 0.050& 0.078 &\textbf{0.100} & 0.050 \\
        LLaMA-2-70B-chat & 0.039& 0.038&0.053 & 0.076&\textbf{0.101}$^\dag$ &0.069 \\
        Mistral-7B-Instruct&0.078&0.070&0.098&0.149&\textbf{0.161} &0.146 \\\hline
    \end{tabular}
    \label{tab:main_res}
\end{table*}
\subsection{Quality Check}
\label{sec:quality-check}
On top of the intermediate quality check stages in the above process (e.g., filtering out inconsistent answers and low-scored scripts, etc.), we finally perform two other rounds of quality checks to further ensure high data quality. 

First, we assess step dependency annotation quantitatively: in three randomized experiments, we sample 100 instances from the ProScript dev and test sets. Following~\citet{sakaguchi-etal-2021-ProScript-partially}, we compare the pair-wise precision, recall, and F1 score of our generated dependency annotations with human performance, which is obtained via asking crowdworkers to annotate partially-ordered scripts for randomly shuffled steps.\footnote{Precision, recall, and F1 score are defined as follows: $\text{Precision} = \frac{|E\cap\hat{E}|}{|E|}$, $\text{Recall} = \frac{|E\cap\hat{E}|}{|\hat{E}|}$, $\text{F$_1$} = \frac{2*\text{Precision}*\text{Recall}}{\text{Precision}+\text{Recall}}$, with \emph{E} being the gold edges and \emph{$\hat{E}$} denoting the predictions in each discourse graph, respectively.} 
Our annotation method has near human-level performance, as reported in Table~\ref{tab:dep_val}.

In addition, we randomly sample 80 instances with a mixture of LLM and human-annotated data and qualitatively survey experts without informing them which data points are human-annotated. We follow the `prescriptive' approach in~\citet{rottger2021two} by instructing them to consider the acceptability of the task time estimations and step ordering constraints. Human-annotated and LLM-generated instances receive similar levels of acceptability.

\section{Benchmarking Experiment}
\label{sec:experiment}
We are interested in answering the following questions. First, can a model efficiently solve asynchronous planning tasks with existing prompting techniques such as \emph{k}-shot~\cite{brown2020language} and Chain of Thought prompting \citep[CoT;][]{wei2022chain}? Second, can we develop a better method to prompt models to improve their performance? Third, how do scale effects manifest when varying problem complexity and model size? Last, is an LLM's performance robust to trivially different linguistic or graphical prompts? We design the experiments accordingly.

\subsection{Experimental Setting and Design}
\label{sec:exp_design}
We conduct experiments with GPT-3.5, GPT-4, and Command, three closed-source LLMs, as well as LLaMA-2-70B-chat and Mistral-7B-Instruct (v0.2), two open-source LLMs.\footnote{In a preliminary experiment, we also tested \mbox{CodeLlama-34B}~\cite{roziere2023code} and \mbox{Phi-2}~\cite{gunasekar2023textbooks} , but we exclude them from the evaluation due to poor performance. Our observation that the code models perform poorly is in line with \citet{la2024code} and \citet{liu2024codemind}, showing that simulation is more difficult than generation.} We first experiment with different language descriptions of our problem in a zero-shot setting with 100 sampled prompts (see details in Appendix \ref{sec:vary-nl-prompt}) and use the best-performing one for the successive experiments.

We then benchmark our models in full scale in four prompting regimes: (i) \textbf{zero-shot}: only prompting models with task descriptions without additional information or training; (ii) \textbf{\emph{k}-shot} \cite{brown2020language}: prompting with \emph{k} in-context instances with desired outputs preceding the task description; (iii) \textbf{zero-shot with CoT} \citep[zero-shot+CoT;][]{kojima2022large}: prompting the model with the task description along with the instruction \texttt{`Let’s think step
by step'}, and (iv) \textbf{\emph{k}-shot with CoT} \citep[\emph{k}-shot+CoT;][]{wei2022chain}: prompting the model with \emph{k} in-context instances with CoT illustrations for the problem-solving process and desired outputs preceding the task description with CoT.\footnote{We use $k=3$. We do not conduct full-scale benchmarking on other prompting techniques such as Chain-of-Thought Self-Consistency \citep[CoT-SC;][]{wang2022self} and Tree of Thought \citep[ToT;][]{yao2024tree} as they primarily use standard IO prompts like CoT, and our method can be deployed in addition to these prompting techniques. We show that our method is superior to using CoT-SC and ToT alone for a more comprehensive comparison in Appendix~\ref{sec:sc_tot}. See also the latency analysis for cost-performance trade-offs in Appendix~\ref{sec:latency}.}

Then, we experiment by sampling 100 instances for the adjacency list, edge list, adjacency matrix, and csr in the setting of PLaG (explicit graph). We use the best type for full-scale PLaG experiments (explicit graph/BaG). Prompt examples are given in Appendix~\ref{sec:prompt-bench}.

\subsection{Experiment Results}\label{sec:exp-results}

We evaluate each model's performance by the accuracy of correctly reporting the shortest time needed for different plans. Main results are in Table~\ref{tab:main_res}.\footnote{If a closed-source model does not return anything due to content filtering, we consider the answer to be false. In Appendix~\ref{sec:exclude_invalids}, we provide analyses where we exclude such invalid instances.}

The strongest performance is obtained by GPT-4 with PLaG (BaG). This is surprising given that GPT-3.5 does better than GPT-4 (though not very well) on the zero-shot, zero-shot+CoT, and \emph{k}-shot settings, which lack explicit illustrations. A solid performance gap divides open-source models from GPT models, although Mistral-7B-Instruct performs better despite being much smaller than LLaMA-2-70B-chat.

PLaG (our method) successfully boosts the performance of all models. This is particularly interesting considering that while natural language prompts in the vanilla setting essentially represent the same information as the graphs, explicitly providing prototypical graph-structured data enhances LLM performance and highlights its inherent limitation of reasoning at a conceptual level. Furthermore, many other real-world tasks in natural language such as dialogue state tracking \cite{lin-etal-2021-knowledge} can be abstracted as graphs. Thus, our finding is relevant for future research on enhancing conceptual representations in LLMs.

Surprisingly, PLaG with BaG, which does not require external processing to supply new graphs explicitly in every task description, improves the performance of the most capable models (GPT-3.5 and GPT-4) across all complexity levels \textbf{off the shelf}.\footnote{We emphasize that the superior performance of BaG does not result from noise in sampling (see Appendix~\ref{sec:plag_noise} for discussion). We hypothesize that the superior performance of 
the explicit graph in other models results from their incapability of generating accurate graph representations.} These results suggest that PLaG benefits LLMs by adding graph information and indicates the potential of boosting capable models' performance in planning without external processors.

\begin{figure}[t]
     \centering
    \includegraphics[width=0.9\linewidth]{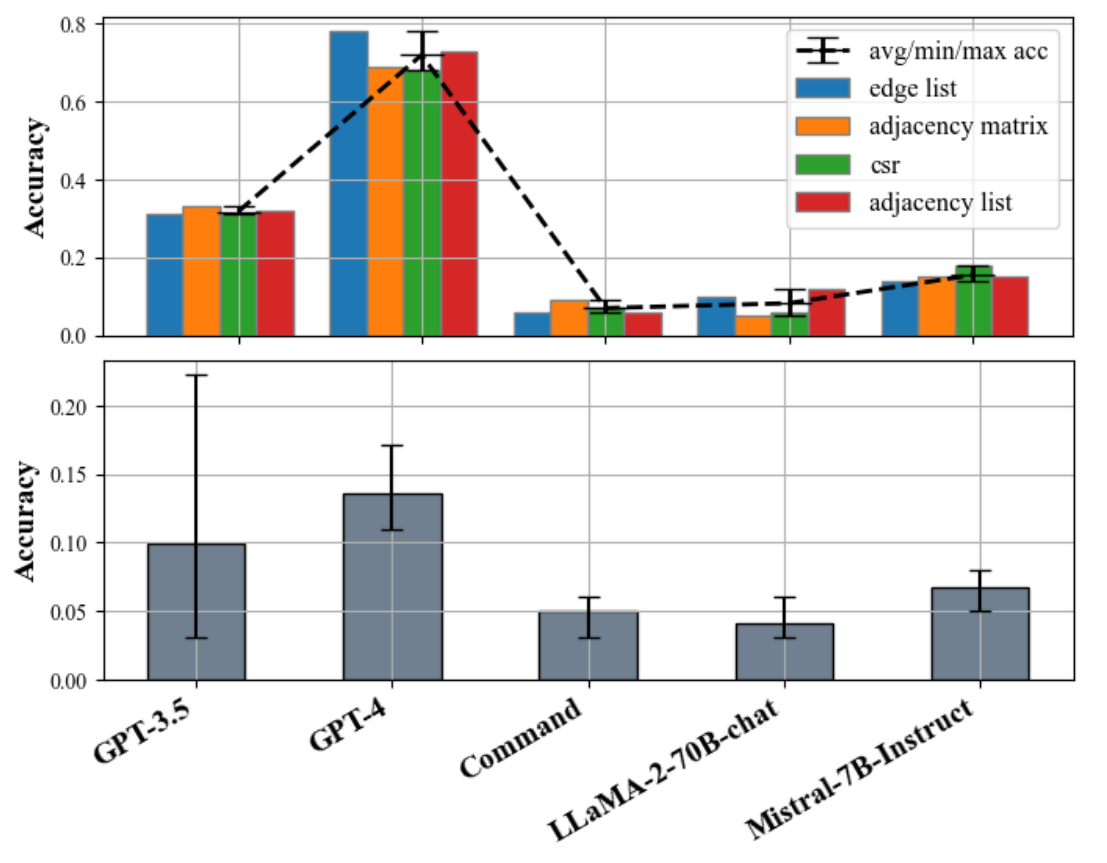}
    \vspace{-5mm}
    \caption{\small The upper plot refers to average model accuracy with 100 instances of PLaG (explicit graph) in different graphs. Colored bars refer to model performance with different graph types. Black dashed lines refer to average accuracy with different graph types. The lower plot refers to the average zero-shot accuracy in 100 instances of different text prompts (without economic usage). Error bars in both plots refer to worst/best performance per model.}
    \label{fig:vary-prompt-graph}
\end{figure}

Next, we report our experiment results on different text prompts and graph types. As shown in Figure~\ref{fig:vary-prompt-graph}, different text prompts and graph types induce variations in model performance, and models have different preferences for these variables. In general, using more succinct and more natural expressions (see Appendix \ref{sec:nl-prompt-app} for details) tends to downgrade model performance (results are reported in Appendix~\ref{sec:economic_exp}), a hint that models cannot adapt to slight variations of the same prompts.

\section{Further Analysis of GPT-3.5/4 Results}
\label{sec:analysis}
This section investigates which factors influence the most potent models in our task, namely GPT-3.5 and GPT-4. We first relate an LLM's accuracy with task complexities, then provide an ablation study to identify the salient characteristics that make a planning problem inherently complex. Next, we use a synthetic dataset that covers and goes beyond the distribution of AsyncHow to estimate model performance in the potential scenarios that might fall out of the distribution of our benchmark. Last, we perform a qualitative analysis to provide further rationales for some surprising phenomena observed in our results.

\subsection{Accuracy vs. Complexity}
\label{sec:acc-complexity}

In Figure~\ref{fig:acc-ve} (Section~\ref{sec:intro}), we plot the accuracy of GPT-3.5 and GPT-4 as a function of the task complexity $|V|+|E|$.\footnote{We show in Appendix~\ref{sec:distr-ood} that $|V|$ and $|E|$ equally contribute to the complexity of a planning task, with no clear dominance of one over the other.} Generally, the accuracy negatively correlates with the complexity of the task for all models and all settings, with graphs of complexity $|V|+|E|\ge 18$ already resulting in challenging problems for the most capable model and best setting. The complexity measure also predicts the model accuracy trends well in settings without graphs. We notice a little jump at complexity $|V|+|E|\ge 20$, for which we will discuss possible reasons in Section~\ref{sec:qual_study}.

Without our method, GPT-4 with \emph{k}-shot + CoT consistently outperforms any other settings by a solid margin, while all the other settings have comparable performances independently from the number of illustrations provided or the model employed (Figure~\ref{fig:acc-ve}, upper).

Our method (PLaG) consistently improves over \emph{k}-shot+CoT, the best method without PLaG, among tasks of all complexities (Figure~\ref{fig:acc-ve}, lower) for both GPT-3.5 and GPT-4. In line with what was observed before, the accuracy drops significantly with complex planning tasks, once again proving that LLMs are not yet robust enough to be deployed as generally intelligent agents in planning. 

\subsection{Ablation Study}
\begin{figure}[t]
    \centering
    \includegraphics[scale=0.5]{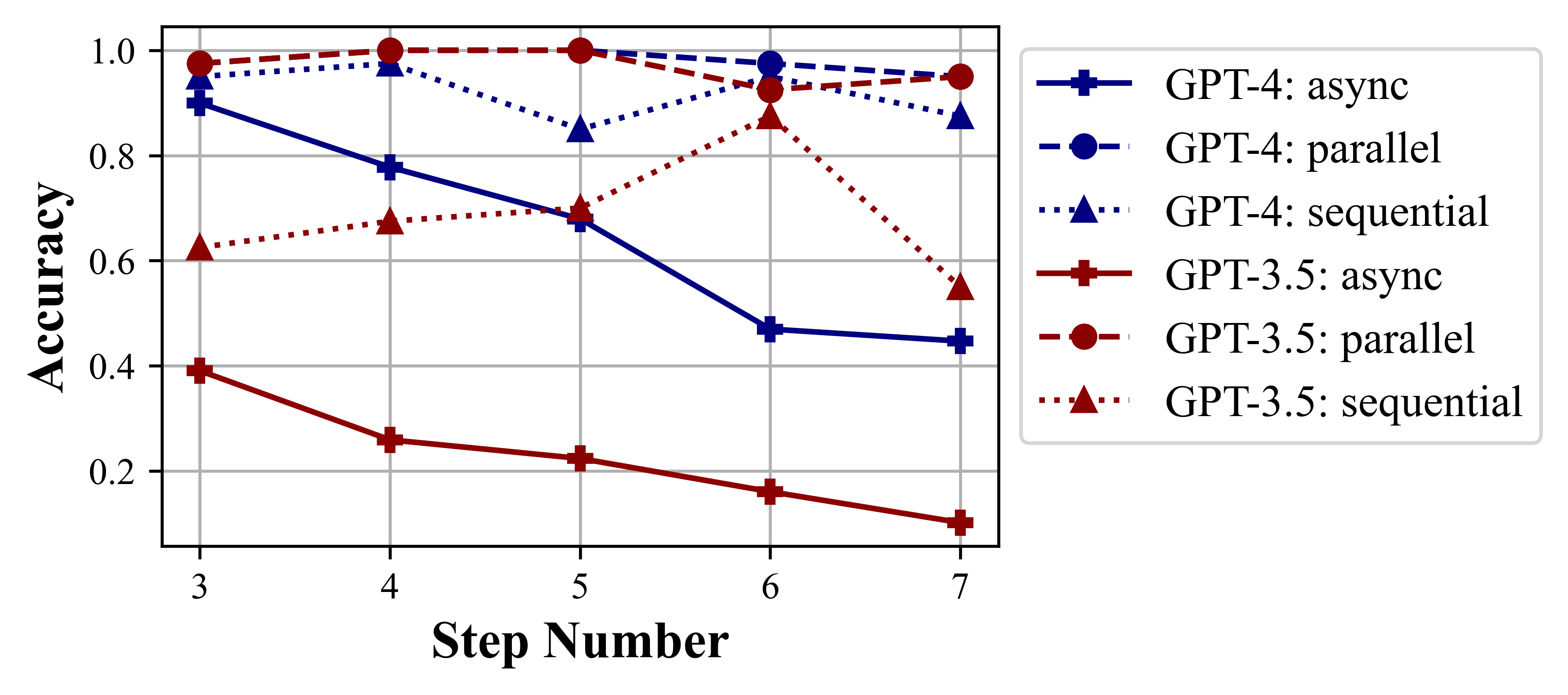}
    \caption{Comparison of parallel/sequential plan execution accuracy with asynchronous plans. All experiments are done in the setting of \emph{k}-shot + CoT. {\color{blue}{Blue}} and {\color{red}{red}} lines refer to GPT-4 and GPT-3.5 results, respectively.}
    \label{fig:baseline}
\end{figure}

Solving our planning task requires a certain degree of compositionality in combining \emph{time comparison}, \emph{time summation}, and \emph{constraint reasoning} correctly. 
We perform an ablation study to identify which skills LLMs lack. We sample 200 sequential planning tasks (i.e., the optimal time calculation only requires summation) and fully parallel tasks (i.e., the optimal time calculation only requires comparison) from the non-asynchronous part in our generated dataset. We sample equal numbers of plans per step with a minimum of three steps and a maximum of seven, as smaller plans are trivial while higher ones are sparse in AsyncHow, which can potentially lead to sampling bias. We experiment with \emph{k}-shot + CoT and compare model performance across step numbers in different plan types in Figure~\ref{fig:baseline}.

While GPT-3.5 and GPT-4 have a similar accuracy in parallel tasks requiring time comparison, GPT-4 outperforms GPT-3.5 on sequential tasks, i.e., at time summation. Our results suggest a performance gap between parallel/sequential and asynchronous plans for both models. We conclude that reasoning about task constraints adds special difficulty on top of time comparison and summation for LLMs. 

\subsection{Out-of-distribution Probing}

We estimate LLMs' performance on out-of-distribution data points whose complexities fall out of the data-rich part (i.e., complexity $|V|+|E| < 20$) of AsyncHow. 
As our naturalistic planning problem can be cast into longest-path graph search (refer to formalism in Section~\ref{sec:formalism}), we generate a synthetic dataset of 2,000 data points evenly distributed between complexity 10 to 40 for \emph{prototypical} shortest-path graph search (we can cast the longest-path search to shortest-path search by negating edge weights). By \emph{prototypical}, we refer to formulating prompts for dynamic programming problems where an LLM is queried to compute the longest path on a graph with numerical edge weights to simulate time durations for each node the edge starts from. For consistency, we sample the graphs as similar to AsyncHow data as possible (see details in Appendix, Section~\ref{sec:distr-ood}).
We prompt GPT-3.5 and GPT-4 in zero-shot + CoT with their respective best graph representations found in Section~\ref{sec:exp-results}. We compare its accuracy with the accuracy in zero-shot + CoT in the naturalistic experiment.

\begin{figure}[t!]
    \centering
    \includegraphics[scale=0.5]{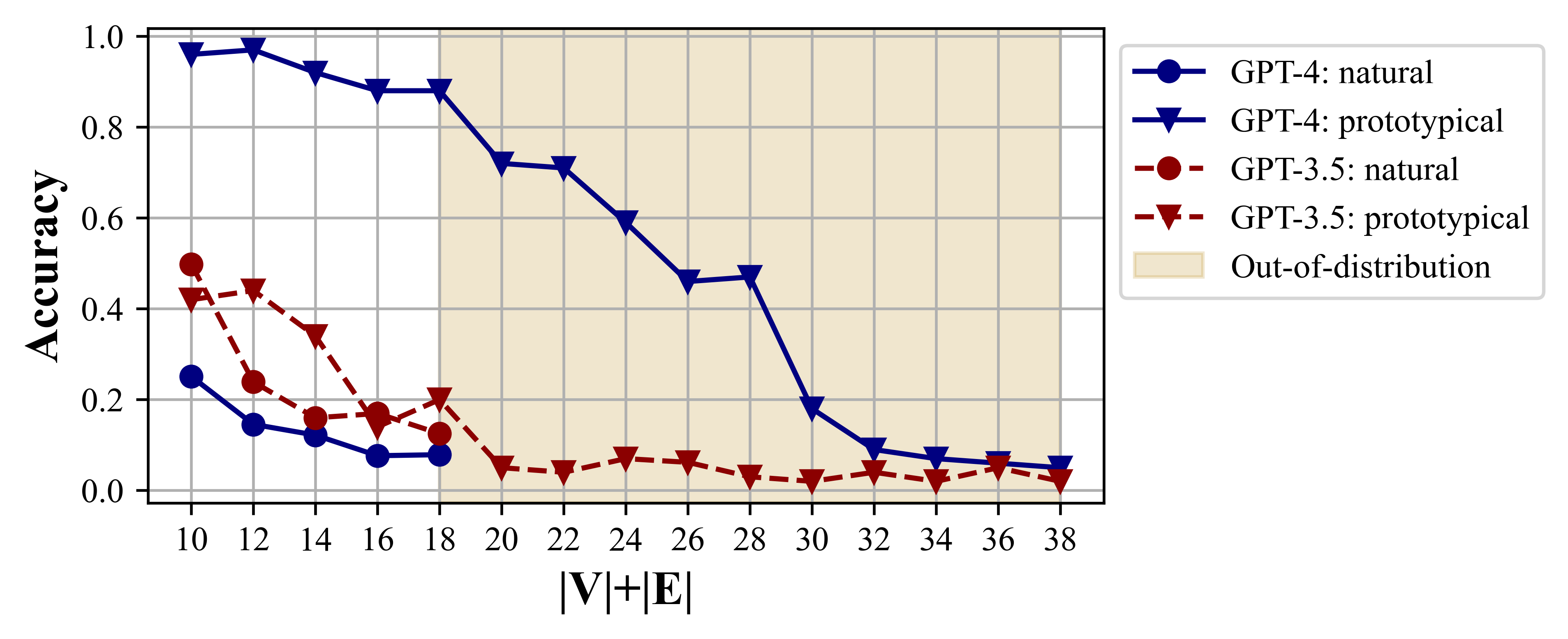}
    \caption{GPT-3.5 and GPT-4 performance on \emph{prototypical} longest path search problem and on \emph{natural} AsyncHow  task with zero-shot + CoT.  Models' respective best graph types found in Section \ref{sec:exp-results} are used in prototypical probing.}
    \label{fig:ood}
\end{figure}

As shown in Figure \ref{fig:ood}, graph search accuracy shows a down-going trend similar to the in-domain naturalistic data, which indicates that model performance in naturalistic planning tasks is likely to follow the pattern of synthetic data and continue to drop with complexity further increasing. This outcome strengthens other findings that LLMs can be unreliable routine simulators~\cite{la2024code}.

Interestingly, although solving essentially the same task, the performance of GPT-4 is much higher in the prototypical setting than the naturalistic one. In comparison, GPT-3.5 derives little benefit from the prototypical setting. We impute this gap to several concurrent factors. First, computing the optimal plan with durations expressed as numbers is easier than naturalistic time conversion (we will discuss this in the next subsection). Second, naturalistic planning requires turning language into an effective procedure, which adds to the difficulty of processing \emph{prototypical} graphs. The results also shed light on the reasons behind the boost of performance caused by PLaG: PLaG points a model to a setting it is already familiar with and better masters.

\subsection{Qualitative Study}
\label{sec:qual_study}
We qualitatively overview some failures and successes of LLMs, which shed light on edge cases that are of interest to understanding their capabilities and limitations.
\paragraph{Wrong answers in easy problems.}
Even for low-complexity planning instances, GPT-4 may incur trivial errors.
It emerges that errors tend to fall into a few macro-categories:  (i) \textbf{parallelism error} where LLMs cannot efficiently parallelize as many steps as possible: e.g., when step 3 (10 min) can be done together with step 1 (5 min) and 2 (15 min), the model only parallelizes 1 and 3 but schedules step 2 to follow them; (ii) \textbf{time unit conversion error} where LLMs cannot efficiently convert time units to common measures for calculation: e.g., 3 weeks and 1 hour is wrongly converted to be 5,041 hours (30 weeks and 1 hour) in the final answer. Our findings are in line with \citet{dziri2023faith} and \citet{la2024code} in that LLMs tend to prefer linear pattern matching and are prone to mistakes in time carries \cite{wang2023tram}.

\paragraph{Correct answers in hard problems.} 
For graphs of complexity $|V|+|E| \geq 20$, GPT-3.5 and GPT-4 perform slightly better than that at $|V|+|E|=18$, which have lower-class complexity. 
We impute this phenomenon to (i) the sparsity of graphs for higher complexities and (ii) an implicit bias of our benchmark towards easier data conversions for more complex planning tasks. See a more in-depth discussion of this phenomenon in Appendix~\ref{sec:time_unit}.

\paragraph{Why is BaG better than explicit graph?} 
A symbolic processor deterministically generates correct graph representations, while BaG prompts a model to generate its internal representation of the problem with no promise of complete correctness. However, among PLaG methods, BaG performs slightly better than explicit graphs (i.e., generated algorithmically).
We sample some instances and find that BaG-generated graphs are of the same format as provided in the \emph{k}-shot prompt, and we thus impute the performance gap to a sub-optimal positioning of the graph in the former setting~\cite{liu2023lost,mao2023prompt}: the explicit graph prompt setting expresses the prompt in the form \texttt{`[Task description with graph] Answer:'}, with the graph appearing in the middle of the context, which can be easily ignored by the model. For BaG, by contrast, the prompt has the form \texttt{`[Task description] Answer: [Graph]'}, where the graph is generated at a successive step and easier for models to take into account.

\section{Related Work}
\label{sec:rw}
\paragraph{LLMs for planning.}
Works focusing on automatically generating plausible plans for daily tasks show that LLMs can be used to develop reasonable and ordered actions or goals~\cite{madaan2022language, xie2023translating, yuan-etal-2023-distilling}. The work most similar to ours is~\citet{sakaguchi-etal-2021-ProScript-partially}: they collected 6.4K ordered plans via crowdsourcing for ProScript. However, the dataset is insufficient to serve as a benchmark for asynchronous planning as relevant data points are sparse and lack diversity. 

Another line of work focuses on finding the optimal plan for domain-oriented tasks such as robotics. Although LLMs can be readily deployed to parse natural language into logical elements, they alone cannot develop optimal plans to accomplish a given goal without external symbolic processors~\cite{collins2022structured,valmeekam2022large,lawless2023want,lin2023text2motion,liu2023llmp,yang-etal-2023-coupling}. While these works focus on domain-oriented tasks using external symbolic processors, we close the gap between structured and naturalistic tasks and show the potential of solely using LLMs for these tasks.

\paragraph{LLMs for graph reasoning.}
Two complementary lines of work inform LLMs with graphs and can be categorized into \emph{implicit} and \emph{explicit} methods. Implicit methods help decompose task goals into atomic steps~\cite{huang2022language, valmeekam2022large, sakib2024consolidating} and help explain complex reasoning processes~\cite{madaan-etal-2021-give, saha-etal-2021-explagraphs, besta2023graph,dziri2023faith}. Works on explicit graphs incorporate external knowledge and reason about more complex problems such as multi-hop question answering~\cite{chen2023exploring, park2023graph,ye2023natural}. Our work shows that instructing LLMs to consider problems like graphs can improve their performance in planning. It also complements recent discoveries suggesting that LLMs' performances negatively correlate with the complexity of graph problems~\cite{fatemi2023talk, guo2023gpt4graph, wang2023can}, showing that the conclusion also holds in relevant naturalistic tasks.

\paragraph{Discourse Representation Theory.}
Humans produce and understand language in a structured way. For instance, when writing a paragraph, people can have a main point and then elaborate on the supporting elements of the discourse~\cite{flower1992planning,limpo2018effects}.
Graphs offer a structured representation of the discourse, with elements as nodes and relations as edges representing elaboration and parallel or temporally/causally linked actions~\cite{wolf2004discourse,presutti2012knowledge,ma2022enhanced}. Recent works suggest that LLMs do not possess identical linguistic representations as humans since they do not compositionally process language and perform tasks in a human-like way \cite{bertolini-etal-2022-testing,press2022measuring,chen2023skills,dziri2023faith}. We find in our work that enriching natural language prompts in a structured manner helps LLMs in relevant tasks.

\section{Conclusion}
\label{sec:conclusion}
In this paper, we automatically generate a benchmark, AsyncHow, and assess LLMs for their performance in asynchronous plan reasoning. We find that if not provided with a detailed illustration of the task solution process, all models behave extremely poorly in our task. We propose a formalism to classify naturalistic asynchronous planning tasks, which successfully predicts LLMs' performance patterns. We propose PLaG, a method that consistently boosts SOTA model performance across all task complexity levels off the shelf. Despite this, we find that model performance still drastically downgrades with increasing task complexity, which calls into question using them as digital devices or generally intelligent agents. 

\section*{Limitations and Future Work}
Some limitations of this work are as follows. We assume that infinite resources are available in our benchmarking, while only finite resources may be available for tasks in real life. Second, we only consider time cost in plan optimization, while realistically, other restrictions, such as preferences, should be considered. Regarding future work, it will be interesting to further elaborate on our benchmark with the proposed techniques and our dataset to add more elements such as resource constraints, multimodality, multilingualism, or other robotics/reinforcement learning features. Practitioners can also scale up the complexity of the benchmark to more complicated tasks. Another promising avenue of research is to compare the performance patterns of LLMs to those of humans (i.e. are LLMs likely to make the same mistakes as humans in asynchronous plan reasoning).

\section*{Data Access Statement}
The dataset used in this paper can be found in \url{https://github.com/fangru-lin/graph-llm-asynchow-plan}.

\section*{Impact Statement}
\label{sec:impact}
This paper presents work whose goal is to advance the field of machine learning. There are many potential societal consequences of our work. It not only unveils the limitations of SOTA LLMs but can also potentially influence many downstream tasks such as job scheduling with the wide application of such technologies. Ethically, since we generate part of our benchmarking dataset from WikiHow, a web data source, users may find relevant content unsafe or uncomfortable. We try our best in the data generation process to leverage the metadata in GPT model outputs to filter out all instances in which either the prompt itself or the reply is flagged as not completely safe concerning hate, self-harm, violence, or sexual aspects. Furthermore, unlike social media for general usage, WikiHow is designed for gaining tips for real-life tasks, which makes it less likely to contain harmful content.

\section*{Acknowledgement}
We thank all the bodies who have provided funding for the authors and for the associated project. FL is supported by Clarendon and Jason Hu studentship. ELM is supported by the Alan Turing Institute. AGC is supported by the Economic and Social Research Council (ESRC) under grant ES/W003473/1, by the Fundamental Research priority area of The Alan Turing Institute, and by the Turing Defence and Security programme through a partnership with the UK government in accordance with the framework agreement between GCHQ and the Alan Turing Institute. JBP is supported by the Engineering and Physical Sciences Research Council (EP/T023333/1). The support of Microsoft under their Accelerating Foundation Models Research initiative in providing Azure credits to AGC is gratefully acknowledged. This work was also supported by compute credits from a Cohere For AI Research Grant to FL. These grants are designed to support academic partners conducting research with the goal of releasing scientific artifacts and data for good projects. Last, we are grateful to the people who offered invaluable feedback and suggestions along the way, and in particular to all reviewers of this paper.

\bibliography{example_paper}
\bibliographystyle{icml2024}

\newpage
\appendix
\onecolumn
\section{Appendix}
\subsection{Extended Preliminaries: Naturalistic Asynchronous Planning}
\label{sec:petri_net}
\subsubsection{Naive Asynchronous Planning with DAG}
Assuming infinite resources (e.g. as many agents and tools as needed to achieve optimal parallelism are available), our task can be considered as finding the optimal time cost $TC$ of a partial-order plan $P$. A partial-order plan $P$ is classically defined as $P= \langle A, O, C \rangle$, where $A$ is a set of actions $a$ (including start and finish), $O$ is a set of ordering constraints which take the form of $a_{i} \prec a_{j}$, meaning $a_{i}$ has to be performed before $a_{j}$, and $C$ as causal links taking the form of $a_{i} \xrightarrow{p} a_{j}$, meaning performing $a_{i}$ meets precondition $p$ needed for $a_{j}$~\cite{alma990119382160107026}.

Specifically for our task, with infinite resources, the formalism of a DAG captures the complexity of finding the optimal order of execution of actions that can or cannot be parallelized in $P$ to minimize time cost $TC(P)$. A DAG $G(P)$ representing $P$ can be defined as~$G(P) = \langle V, E, w \rangle$, where $V$ is a set of nodes, each representing an action $a$ in the planning problem. We serve two auxiliary nodes $v_{\text{src}}$ (START) and $v_{\text{dst}}$ (END) that connect respectively each initial and final component in $G$ on top of other meaningful nodes but do not impact the optimal solution of the problem (as sketched in Figure~\ref{fig:dag_task_desc}). $E$ is a directed set of flow relations representing ordering constraints $O$, while $w$ is a function that assigns weight to all edges in the graph $w: E \rightarrow \mathbb{R}^+$. Each flow relation is associated with a positive number (the \emph{weight} of a connection), namely $w(e_{i,j})$, to express that node/action $v_{i}$ is connected to node/action $v_{j}$ and requires $w(e_{i,j})$ time to be completed. The edges also represent causal links $C$ in that the precondition $p$ for an action/node $a$ is met if all actions/nodes linked to and preceding $a$ are performed. For simplicity, we denote $G(P)$ as $G$ in the remaining part of the paper.

In this setting, finding the time cost for an optimal plan  $P^*$ in a planning problem is equivalent to finding the longest path $G^*$ on $G$ and can be cast as the following optimization problem on a subgraph $G'= \langle V', E', w\rangle, \ G' \subseteq G$:
\begin{equation}
\begin{aligned}
P^* \in \argmin_{P=\langle A,O,C\rangle} TC(P)\\
\Leftrightarrow G^* \in \argmax_{G'=\langle V', E', w\rangle \subseteq G} \sum_{e_{i,j} \in E'}w(e_{i,j})\\
\textrm{s.t.} \ \forall v'_i \in (V' \setminus v_{dst}), \exists! \  v'_j \in v'_{i.\text{next}} \ | \ (v'_j, e'_{i,j}) \in (V',E')\\
(v_{src}, v_{dst}) \in V'. \\
\end{aligned}
\end{equation}
In this formulation, $v'_{i.\text{next}}$ are vertices in $G$ that are successors of $v'_{i}$.

Consider the example in Figure~\ref{fig:dag_task_desc}, where we sketch the DAG to solve the planning task of Figure~\ref{fig:task_desc}. The maximum time to complete the whole task is $65$ minutes (i.e. sequentially executing all actions), while parallelizing all actions violates constraints on the preconditions of some actions (e.g. \texttt{`Roll dough'} and \texttt{`Add filling'} cannot be done simultaneously). Parallelizing \texttt{`Roll dough'}, and \texttt{`Preheat oven'}, and then executing the other actions allows solving the problem optimally. 

\subsubsection{Asynchronous Planning with Petri Net}
\begin{figure*}[h!]
    \centering
    \includegraphics[width=0.8\linewidth]{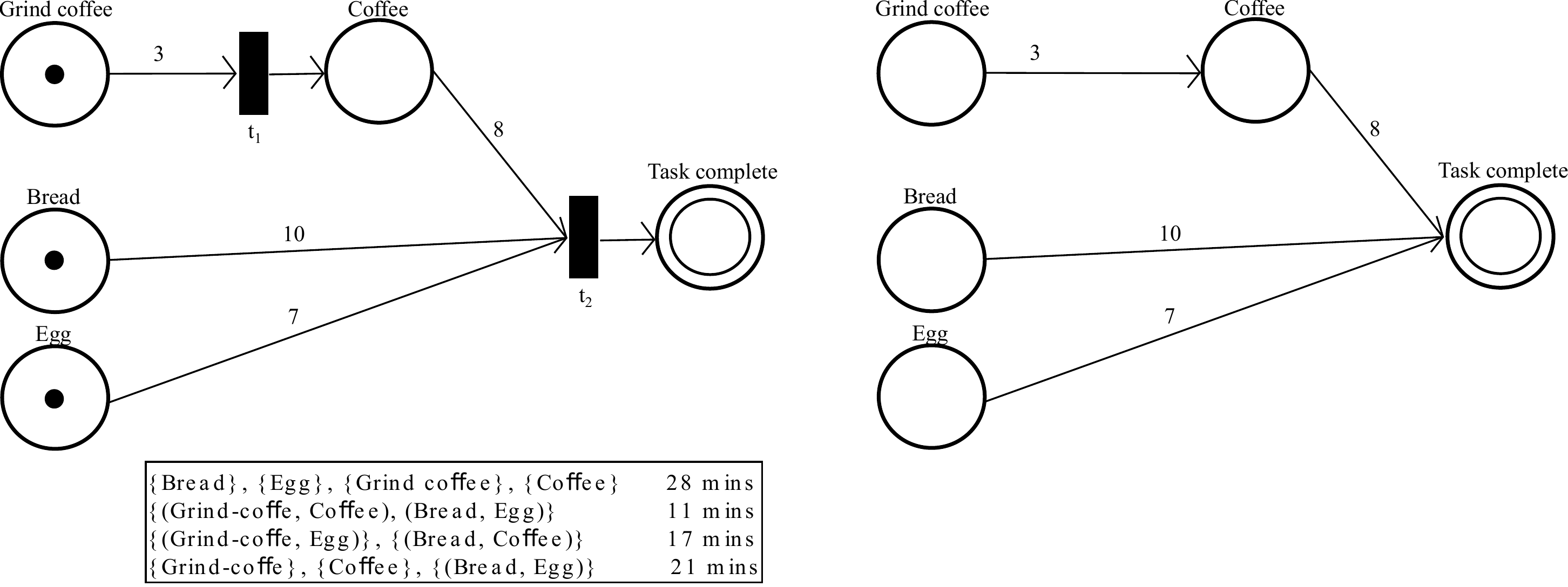}
    \caption{On the left, a Petri net representation of the example of making breakfast. A few admissible \emph{runs} are reported at the bottom with their completion time. Actions executed sequentially (in parallel) are reported in curly (round) brackets. On the right, the DAG used to solve the optimal planning problem. 
    }
    \label{fig:petri}
\end{figure*}

While we assume infinite resources to complete a planning task, the natural extension to the case of finite resources (i.e., not all independent actions can be parallelized) is better captured by the formalism of a Petri net. Consider a task where one needs to make breakfast by grinding coffee (\texttt{`Grind-coffee'}, 3 min), boiling coffee (\texttt{`Coffee'}, 8 min), making toast (\texttt{`Bread'}, 10 min), and frying an egg (\texttt{`Egg'}, 7 min). The problem of finding the optimal order of execution of actions that can or cannot be parallelized is fully captured by the formalism of Petri nets~\cite{petri1962kommunikation}. A Petri net consists of a tuple~$N = (P, T, F)$, where $P$ and $T$ are disjoint finite sets of places and transitions, and $F$ is a directed set of flow relations associated with a positive number (the \emph{weight} of a connection), namely $(i, j, f_{i,j} \ge 0)$. For an initial configuration of $N$, an action $p \in P$ `fires', sequentially or simultaneously with other actions, a transaction $t \in T$, if it contains a token, usually represented as a circle that encompasses a single dot.\footnote{We assume states can have at most one token and that a transaction is activated if each `firing' state contains a token.} An optimal planning problem is equivalent to finding the shortest transition in a Petri net. Consider the example in Figure~\ref{fig:petri}: while one can execute the actions \{\texttt{`Grind-coffee'}, \texttt{`Bread'}, \texttt{`Egg'}\} simultaneously or in parallel, with the expected completion time reported on the edge, the action \texttt{`Coffee'} must follow \texttt{`Grind-coffee'}. On the other hand, one can execute the actions \{\texttt{`Egg'}, \texttt{`Grind-coffe'}\} in parallel and start the action \texttt{`Coffee'} when \texttt{`Grind coffee'} is complete. Similarly to Figure~\ref{fig:task_desc}, the minimum amount of time required for the task is $28$ minutes, while parallelising \{\texttt{`Grind-coffee'}, \texttt{`Bread'}, \texttt{`Egg'}\} and then execute \texttt{`Coffee'} when \texttt{`Grind-coffee'} is complete allows solving the problem optimally. For a Petri net $N = (P, T, F)$ representing a planning problem, the maximum completion time is the sum of the weights on the longest path (while the minimum is the minimum time to reach the last transition from any parent node).

\subsubsection{On the Complexity of a Planning Task with Finite Resources}
With infinite resources, i.e., assuming one token per initial node (and thus can `fire' the transition), finding the optimal plan is equivalent to finding the longest path from the initial to the final state and can thus be computed efficiently.
A directed acyclic graph (DAG) is obtainable from a Petri net by discarding the set of transitions and reversing the sign on each edge, then using a search algorithm that is linear in complexity for series-parallel graphs.
Formally, a DAG for a planning task has the following formulation: $G=\langle P, W \rangle$, such that $w_{i,j} = -f_{i,j}, (f_{i,j}, w_{i,j}) \in (F,W)$, where $P$ is the set of states and $F$ the set of transitions and weights in the correspondent Petri net. Figure~\ref{fig:petri} (right) reports the DAG for making breakfast.
On the other hand, when resources are finite, finding the optimal planning corresponds to an optimization problem that is generally NP-hard to solve~\cite{graham1979optimization,jain1999deterministic} as one has to estimate the number of reachable states (the state space) from a combinatorial number of initial configurations with $k$ resources. While the exact size of the state space depends on the constraints on sequential actions, such a number is combinatorial and upper-bounded by $2^{n(n-1)}$, where $n$ is the number of actions and equivalently the number of states in the correspondent Petri net. Combinatorial optimization algorithms, including genetic algorithms and simulated annealing, are usually applied to search for solutions in NP-hard problems as such since exact methods tend to be too complex.

\subsection{Data Generation Details}
\label{sec:data-gen-app}
\begin{figure*}[h!]
    \centering
    \includegraphics[scale=0.38]{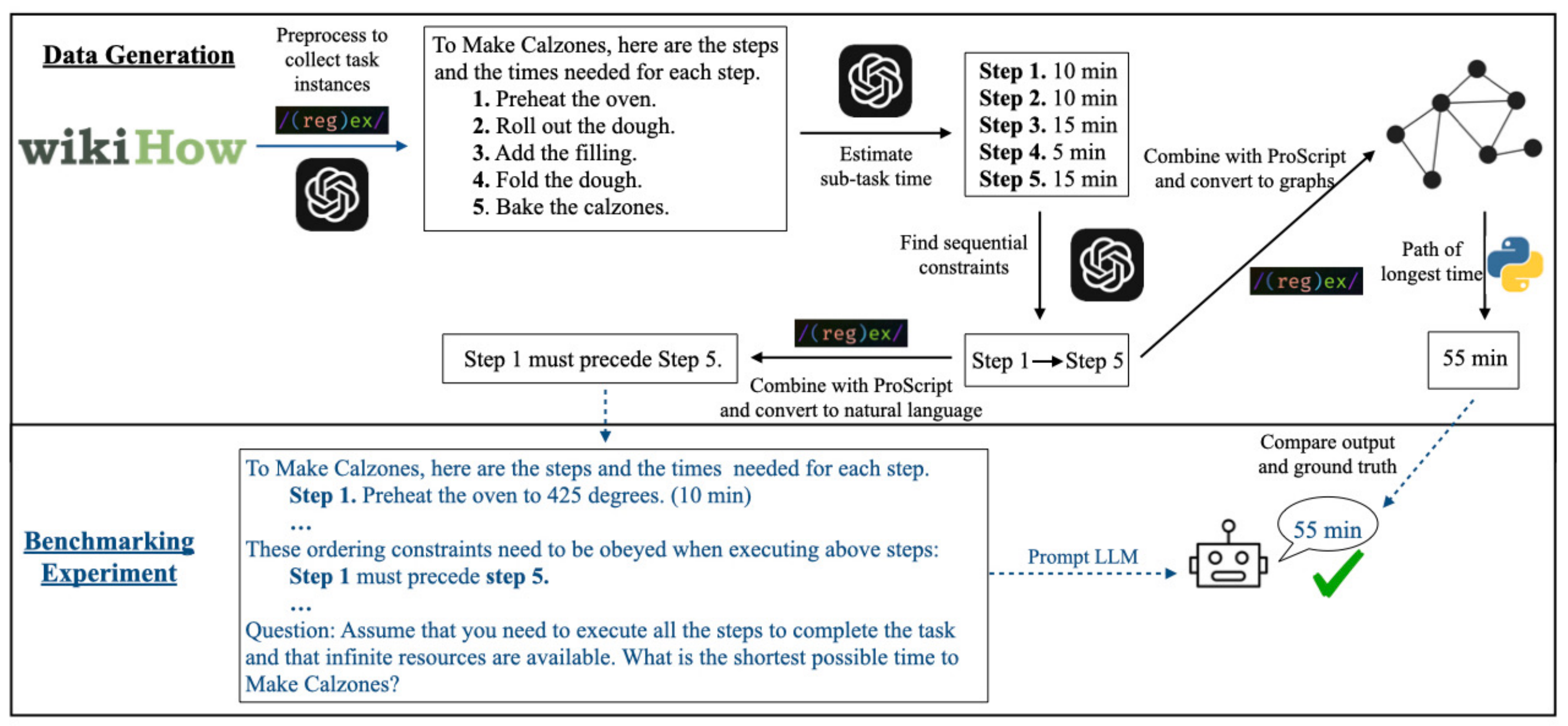}
    \caption{Workflow of data generation and experiment setting. During data generation, we first conduct \textbf{preprocessing} to collect high-quality scripts from WikiHow that suit our goal with regular expression and GPT-3.5, then prompt GPT-3.5 to \textbf{generate time estimations}. Data points with non-numerical estimations are discarded. The remaining data points are run through GPT-4 to \textbf{get step dependencies} in dot language. Then, we combine our dataset with ProScript and pull together information to \textbf{generate natural language prompts}. We use an external oracle symbolic processor to \textbf{generate graphs} representing given information as well as \textbf{generate ground truth answers} for optimal task time duration. In benchmarking experiment, we prompt LLMs with natural language task descriptions and compare models' answers with gold symbolic processor outputs.}
    \label{fig:workflow}
\end{figure*}
In addition to existing data in ProScript\cite{sakaguchi-etal-2021-ProScript-partially}, we use WikiHow~\cite{koupaee2018wikihow,zhang-etal-2020-reasoning} as a base dataset to derive the planning tasks we need. The data generation strategy is schematized in Figure \ref{fig:workflow}, which consists of five steps: \textbf{preprocessing}, \textbf{time duration annotation}, \textbf{step ordering constraint annotation}, \textbf{natural language prompt generation}, and \textbf{graph and gold answer generation}. 
We first leverage GPT-3.5/4 together with keyword-catching algorithms to filter out low-quality examples (Section \ref{sec:preprocess}) and estimate the time duration of planning steps (Section \ref{sec:time-annotation}), and their dependencies for unannotated WikiHow data (Section \ref{sec:step-ordering}). Then we combine the LLM-annotated WikiHow data (about 1\emph{k}) with qualified human-annotated data in ProScript (about 0.6\emph{k}), to generate natural language prompts based on pre-defined templates (Section \ref{sec:nl-prompt-app}) and optimal planning time deterministically by Python, which is by construction correct (Section \ref{sec:graph_gen}). We perform automatic and human data quality validation and show that our dataset is similar in quality to end-to-end human-annotated data (Section~\ref{sec:quality-check}). This process culminates with AsyncHow, a curated list of 1.6K data points for planning, which enables benchmarking models for asynchronous planning against the gold answer provided by symbolic processors.

We now provide details about the data generation process. WikiHow dataset consists of different script types, namely (i) flat scripts, with solely the task the name and step descriptions; (ii) multi-method scripts, which describe different methods to solve one task; and (iii) multi-part scripts, which describe complementary parts to solve a task. We exclude script type (i) and (iii) instances entirely if any of their sub-parts do not meet our requirement; we only exclude the unqualified methods in script type (ii) as we can view different methods as different independent scripts. We clarify the qualification criteria below.

\subsubsection{Pre-processing}
\label{sec:preprocess}
First, we take the scripts with a collaborative rating score (i.e., the helpful percentage in WikiHow) higher than $60\%$ to retain only high-quality plans marked as useful by the users of WikiHow. We leverage GPT-3.5 and keywords methods to filter out plans with optional steps (e.g., \texttt{`If you are not happy with the results, proceed to follow steps'}) and others that do not fit into our research goal (we report a list of filtering keywords and descriptions in Appendix \ref{sec:exclude_words}). Inspired by~\citet{wang2022self}, plans that contain unnecessary steps are further filtered with GPT-3.5 in a few-shot setting by sampling three answers (with the temperature set to one) to exclude those in which the majority vote does not agree on all parts/steps being necessary for a script (see prompts and details in Appendix \ref{sec:prompt_generation}).

\subsubsection{Time Annotation}
\label{sec:time-annotation}
We use GPT-3.5 to estimate the time duration per step. We use a zero-shot setting by sampling three answers (with temperature set to one) and exclude instances whose steps duration an LLM cannot quantify numerically (see prompts and details in Appendix \ref{sec:prompt_generation}). Empirically, we keep the longest among all the time estimations GPT-3.5 proposes for each step. We note that the longest time estimation is not necessarily always the only acceptable answer. We do not particularly verify time annotation because the task time estimation in a less grounded setting as ours tends not to have a unique gold answer (e.g. \texttt{`finding a gym'} may take five minutes or a week to different people), and GPT-3.5 is reported being a reliable annotator for this task~\cite{jain2023language}. Furthermore, our interest is in assessing whether an LLM outputs the optimal plan for a task, and we expect end users to supply different time durations when querying a model.

\subsubsection{Step Ordering Constraint Annotation} 
\label{sec:step-ordering}
We use GPT-4 in zero shot with the temperature set to one to sample five answers per prompt to annotate dependencies among steps (see more details in Appendix \ref{sec:prompt_generation}). Specifically, we first shuffle steps in each script and then use \emph{dot} language, an unambiguous syntax (e.g., \texttt{`step 1 must precede step 2'} is expressed as \texttt{`step 1 $\rightarrow$ step 2'}), to obtain step dependencies to be compliant with the ProScript format. 
For flat scripts type (i), we annotate dependencies among all steps, while for multi-method scripts type (ii), we annotate dependencies by method; multi-part task dependencies in type (iii) are annotated first among different parts and then among different steps per part, which are then combined to formulate the final dependencies. 
After removing redundant dependencies, we keep data points that have at least four consistent answers that form asynchronous plans and discard the others (e.g. in an answer saying \texttt{`step 1 $\rightarrow$ step 2', `step 2 $\rightarrow$ step 3', `step 1 $\rightarrow$ step 3'}, we remove \texttt{`step 1 $\rightarrow$ step 3'}) in a similar vein to self-consistency prompting \cite{narangself}.

\subsubsection{Natural Language Prompt Generation}
\label{sec:nl-prompt-app}
We combine all asynchronous instances with complete time annotation for all meaningful steps in ProScript with our generated asynchronous instances from WikiHow. We filter out all instances with unannotated preparation steps in ProScript or those flagged as unsafe by either GPT-3.5 or GPT-4 in WikiHow during the above generation process. By keeping only high-quality asynchronous plans from the WikiHow and the ProScript datasets, we obtain a collection of 1.6K instances. 

We then generate natural language prompts based on the task information. 
Dot language, which we used for ordering constraint generation, provides an unambiguous syntax to formulate a planning problem, yet users tend to use natural language descriptions to specify such a task. We prompt GPT-4 to express the dependency constraint with different linguistic formulations, and we end up with ten plausible templates. We report all templates in Appendix~\ref{sec:natural_dep} and include prompts for them in our final dataset. 

We further note that people can combine different constraint expressions for succinct usage, a phenomenon widely accepted linguistically known as the principle of quantity/economy~\cite{grice1975logic, vicentini2003economy}. For example, the constraint \texttt{`Step 1 must precede step 2, step 1 must precede step 3'} can be similarly uttered as \texttt{`Step 1 must precede step 2 and 3'}. We, therefore, also include economic usage for these templates by combining the steps following the common preceding step as exemplified above to allow studies about LLMs' robustness to trivially different natural language prompts.

\subsubsection{Graph and Gold Answer Generation}
\label{sec:graph_gen}
To generate gold answers for a planning task, we parse step dependencies with regular expressions and generate an equivalent DAG representing the workflow. 
We generate the optimal time duration for a plan by iterating every sequential path and choosing the longest one (the longest path algorithm would produce an equivalent gold label).
Each planning task is eventually coupled by four types of graph representations: the adjacency and the edge list, the adjacency matrix, and the compressed sparse row (csr). 
Such representations can be used to aid LLMs in structural reasoning and assess LLMs' robustness against different representations of the same graphs.

\subsection{Topic Assignment}
\label{sec:topic-assignment}
WikiHow data is coupled with metadata, including category hierarchy. For the WikiHow proportion of our benchmark, we take the top-ranked category for each instance as its topic. To assign topics for datapoints in ProScript, we use fast-text static embedding \cite{mikolov-etal-2018-advances} trained on 600B Common Crawl data to embed task descriptions for both WikiHow and ProScript data in our benchmark and mean-pooling the representations after removing stop words. We assign a topic its vector by mean-pool all WikiHow task representations for prompts associated with it. We then calculate the cosine similarity and select the highest one as the corresponding category between each topic vector and task vector for each task in ProScript and assign the topic with highest similarity for each task.

\subsection{Keywords Excluded From WikiHow Dataset}
\label{sec:exclude_words}
We exclude instances that contain the following keywords that fall into categories out of our benchmark goal during dataset pre-processing:

As our task is represented without context such as images in the WikiHow webpage, we exclude \textbf{context-dependent words}: \texttt{this, above, below}.\\
We only maintain tasks that can be calculated for their exact time duration so we exclude \textbf{words indicating ongoing process or no time duration}: \texttt{keep, know, knowing, become, be, stay, repeat}.\\
We assume all steps in a plan are compulsory so we exclude \textbf{words indicating optional procedures}: \texttt{opt, if}.\\
We want all steps in a plan to not overlap each other so we exclude \textbf{words indicating parallel constraints}: \texttt{when, while}.\\
We assume steps to have no intervals among them so we exclude \textbf{words indicating intervals between steps}: \texttt{after, before}.

\subsection{Prompts and Settings for Dataset Generation}
\label{sec:prompt_generation}

\begin{table}[h!]
    \caption{Prompts and settings used to generate AsyncHow dataset.}
    \centering
    \resizebox{\linewidth}{!}{%
    \begin{tabular}{c|c|c|c|c|c|c}
        & Task Prompt & N-shot Example & Shot number& System Prompt & Temperature&Sampling strategy \\\hline
        \makecell{Necessity\\Check} & \multicolumn{1}{p{4cm}|}{Here is a script in [TASK]. [TASK DESCRIPTION] Question: Is this script showing different alternatives to complete this task? Let's think step by step then provide final answer yes or no in double quotes. Answer: }&\multicolumn{1}{p{5cm}|}{To 'Make a Chicken Sandwich', here is a script in 'Making a Fried Chicken Sandwich'.
step1: Done!; step2: Add oil to a large frying pan.; step3: Cut the chicken into thin strips and add toppings of your choice.; step4: Get the necessary ingredients.; step5: Mix the batter.; step6: Batter the chicken.; step7: Put each piece of chicken in the pan.
Question: Is this script showing different alternatives to complete this task? Let's think step by step then provide final answer yes or no in double quotes.
Answer: The steps as presented are not in a logical sequential order. However, they don't provide alternative methods to make a fried chicken sandwich but rather are parts of a single method that are out of order. To properly make a sandwich, these steps need to be rearranged into a sensible sequence (e.g., gathering ingredients, preparing the chicken and batter, frying the chicken, and assembling the sandwich).}&5&You are a helpful plan organizer.&1&\multicolumn{1}{p{3cm}}{Sample 3 answers and take the majority vote.}\\\hline
\makecell{Time\\Annotation}&\multicolumn{1}{p{4cm}|}{Here is a script to [TASK]. 
[TASK DESCRIPTION]
Question: Can you estimate each step in this script for a numerical time duration where possible? Let's think step by step and finally answer in a json format {"step index": "duration"}, each entry a line with double quotes and no description}&NONE&0&You are a helpful plan organizer.&1&\multicolumn{1}{p{3cm}}{Sample 3 answers and take the longest time if any. Exclude instances which have unannotatable steps.}\\\hline
\makecell{Step\\Dependency\\Annotation}&\multicolumn{1}{p{4cm}|}{Here are randomly ordered steps needed to [TASK]. 
[TASK DESCRIPTION]
Assume infinite resources are available and that steps should be parallelized where possible. For each step, does it logically need to follow others considering the nature of the task? Let’s think step by step then finally answer in dot language for all necessary constraints, each constraint per line in the format of "preceding step index" $->$ "following step index"}&NONE&0&You are ChatGPT.&1&\multicolumn{1}{p{3cm}}{Sample 5 answers and take the 4 consistent answers after removing auxiliary dependencies. Exclude instances that have less than 4 consistent answers.}\\
    \end{tabular}}
    \label{tab:prompt_generation}
\end{table}

\newpage
\subsection{Natural Language Prompts for Step Ordering Constraints}
\label{sec:natural_dep}

We prompt GPT-4 web interface to provide 10 different ways to express dot language constraints with natural language, and randomly sample from them to form our prompts.

\colorbox{lightorange}{\textbf{Prompt}} 

Suppose you need to execute several steps to complete a task. With dot language, you can represent temporal constraints of the steps such as "preceding step" $->$ "following step". How can you succinctly express the constraint in natural language? Give me 10 different expressions.

\colorbox{lightorange}{\textbf{Response}}

To express the temporal constraints between a preceding step and a following step in natural language, you could say:

"Before starting [following step], complete [preceding step]."\\
"[Preceding step] must precede [following step]."\\
"Upon completing [preceding step], proceed to [following step]."\\
"After [preceding step], [following step] should commence."\\
"Prioritize [preceding step] before advancing to [following step]."\\
"[Preceding step] is a prerequisite for [following step]."\\
"Initiate [following step] subsequent to [preceding step]."\\
"Ensure [preceding step] is done before [following step]."\\
"Sequence the tasks: firstly [preceding step], then [following step]."\\
"[Following step] follows the completion of [preceding step]."

\subsection{Varying Natural Language Prompting Details}
\label{sec:vary-nl-prompt}
We first sample 100 prompts for all templates in \ref{sec:natural_dep} to compute model accuracies in these templates \textbf{without} succinct usage (e.g. \texttt{`Step 1 must precede step 2, step 1 must precede step 3'}). Then, as we note that the succinct usage of such linguistic utterances is widely accepted as the principle of economy/quantity: do not say more than what's needed \cite{grice1975logic, vicentini2003economy}, we choose the best template and test whether a model can achieve better performance \textbf{with} succinct usage (e.g. \texttt{`Step 1 must precede step 3'}). Then, we choose the best-performing prompting setting for the rest of the experiment.

\subsection{Prompt Examples for Benchmarking Experiment}\label{sec:prompt-bench}
\colorbox{lightorange}{\textbf{Zero shot (zero-shot)}}

To create a video game, here are the steps and the times needed for each step.\\Step 1. Learn the basics of programming (180 days)\\Step 2. Learn to use a language that is used in games (60 days)\\Step 3. Learn to use an existing game engine (30 days)\\Step 4. Program the game (90 days)\\Step 5. Test the game (30 days)\\\\\\These ordering constraints need to be obeyed when executing above steps:\\Step 1 must precede step 2.\\Step 1 must precede step 3.\\Step 2 must precede step 4.\\Step 3 must precede step 4.\\Step 4 must precede step 5.\\\\\\Question: Assume that you need to execute all the steps to complete the task and that infinite resources are available. What is the shortest possible time to create a video game? Answer the time in double quotes.\\Answer:

\colorbox{lightorange}{\textbf{Zero shot+CoT (zero-shot+CoT)}}

To create a video game, here are the steps and the times needed for each step.\\Step 1. Learn the basics of programming (180 days)\\Step 2. Learn to use a language that is used in games (60 days)\\Step 3. Learn to use an existing game engine (30 days)\\Step 4. Program the game (90 days)\\Step 5. Test the game (30 days)\\\\\\These ordering constraints need to be obeyed when executing above steps:\\Step 1 must precede step 2.\\Step 1 must precede step 3.\\Step 2 must precede step 4.\\Step 3 must precede step 4.\\Step 4 must precede step 5.\\\\\\Question: Assume that you need to execute all the steps to complete the task and that infinite resources are available. What is the shortest possible time to create a video game? Let's think step by step and then answer the time in double quotes.\\Answer:

\colorbox{lightorange}{\textbf{\emph{k}-shot}}

\#\#\#Examples:\\To Make Calzones, here are the steps and the times needed for each step.\\Step 1. Preheat the oven to 425 degrees. (10 min)\\Step 2. Roll out the dough. (10 min)\\Step 3. Add the filling. (15 min)\\Step 4. Fold and pinch the dough. (5 min)\\Step 5. Bake the calzones. (25 min)\\\\\\These ordering constraints need to be obeyed when executing above steps:\\Step 1 must precede step 5.\\Step 2 must precede step 3.\\Step 3 must precede step 4.\\Step 4 must precede step 5.\\\\Question: Assume that you need to execute all the steps to complete the task and that infinite resources are available. What is the shortest possible time to Make Calzones? Answer the time in double quotes.\\Answer: The shortest possible time to Make Calzones is "55 min".\\...\textbf{[TWO MORE EXAMPLES]}...\\\#\#\#\\\textbf{[ZERO SHOT PROMPT]}

\colorbox{lightorange}{\textbf{\emph{k}-shot+CoT}}

\#\#\#Examples:\\To Make Calzones, here are the steps and the times needed for each step.\\Step 1. Preheat the oven to 425 degrees. (10 min)\\Step 2. Roll out the dough. (10 min)\\Step 3. Add the filling. (15 min)\\Step 4. Fold and pinch the dough. (5 min)\\Step 5. Bake the calzones. (25 min)\\\\\\These ordering constraints need to be obeyed when executing above steps:\\Step 1 must precede step 5.\\Step 2 must precede step 3.\\Step 3 must precede step 4.\\Step 4 must precede step 5.\\\\Question: Assume that you need to execute all the steps to complete the task and that infinite resources are available. What is the shortest possible time to Make Calzones? Answer the time in double quotes.\\Answer: Since step 1 must precede step 5, step 2 must precede step 3, step 3 must precede step 4, step 4 must precede step 5, we can conclude that we must execute step 2, step 3, step 4, then step 5 sequentially, and since step 1 happens before step 5, it can be done in parallel with step 2, 3, and 4, preceding step 5. Since sequentially executing step 2, 3, 4, and 5 takes 10 + 15 + 5 + 25 = 55 min, while sequentially executing step 1 then step 5 only takes 10 + 25 = 35 min, the shortest possible time to Make Calzones is "55 min".
\\...\textbf{[TWO MORE EXAMPLES]}...\\\#\#\#\\\textbf{[ZERO SHOT+COT PROMPT]}

\colorbox{lightorange}{\textbf{PLaG (explicit graph, graph is adjacency list)}}

\#\#\#Examples:\\To Make Calzones, here are the steps and the times needed for each step.\\Step 1. Preheat the oven to 425 degrees. (10 min)\\Step 2. Roll out the dough. (10 min)\\Step 3. Add the filling. (15 min)\\Step 4. Fold and pinch the dough. (5 min)\\Step 5. Bake the calzones. (25 min)\\\\\\These ordering constraints need to be obeyed when executing above steps:\\Step 1 must precede step 5.\\Step 2 must precede step 3.\\Step 3 must precede step 4.\\Step 4 must precede step 5.\\\\Here is the adjacency list representation of the step ordering constraints:\\\{'1': ['5'], '2': ['3'], '3': ['4'], '4': ['5'], '5': ['END'], 'END': [], 'START': ['1', '2']\}\\Time for each step can be represented as a dictionary:\\\{'1': '10 min', '2': '10 min', '3': '15 min', '4': '5 min', '5': '25 min'\}\\\\Question: Assume that you need to execute all the steps to complete the task and that infinite resources are available. What is the shortest possible time to Make Calzones? Answer the time in double quotes.\\Answer: Since step 1 must precede step 5, step 2 must precede step 3, step 3 must precede step 4, \\step 4 must precede step 5, we can conclude that we must execute step 2, step 3, step 4, then step 5 sequentially, and since step 1 happens before step 5, it can be done in parallel with step 2, 3, and 4, preceding step 5. Since sequentially executing step 2, 3, 4, and 5 takes 10 + 15 + 5 + 25 = 55 min, while sequentially executing step 1 then step 5 only takes 10 + 25 = 35 min, the shortest possible time to Make Calzones is "55 min".\\...\textbf{[TWO MORE EXAMPLES]}...\\\#\#\#\\To create a video game, here are the steps and the times needed for each step.\\Step 1. Learn the basics of programming (180 days)\\Step 2. Learn to use a language that is used in games (60 days)\\Step 3. Learn to use an existing game engine (30 days)\\Step 4. Program the game (90 days)\\Step 5. Test the game (30 days)\\\\\\These ordering constraints need to be obeyed when executing above steps:\\Step 1 must precede step 2.\\Step 1 must precede step 3.\\Step 2 must precede step 4.\\Step 3 must precede step 4.\\Step 4 must precede step 5.\\\\\\Here is the adjacency list representation of the step ordering constraints:\\\{'1': ['2', '3'], '2': ['4'], '3': ['4'], '4': ['5'], '5': ['END'], 'END': [], 'START': ['1']\}\\Time for each step can be represented as a dictionary:\\\{'1': '180 days', '5': '30 days', '2': '60 days', '4': '90 days', '3': '30 days'\}\\\\\\Question: Assume that you need to execute all the steps to complete the task and that infinite resources are available. What is the shortest possible time to create a video game? Let's think step by step and then answer the time in double quotes.\\Answer:

\colorbox{lightorange}{\textbf{PLaG (BaG, graph is adjacency list)}}

\#\#\#Examples:\\To Make Calzones, here are the steps and the times needed for each step.\\Step 1. Preheat the oven to 425 degrees. (10 min)\\Step 2. Roll out the dough. (10 min)\\Step 3. Add the filling. (15 min)\\Step 4. Fold and pinch the dough. (5 min)\\Step 5. Bake the calzones. (25 min)\\\\\\These ordering constraints need to be obeyed when executing above steps:\\Step 1 must precede step 5.\\Step 2 must precede step 3.\\Step 3 must precede step 4.\\Step 4 must precede step 5.\\\\Question: Assume that you need to execute all the steps to complete the task and that infinite resources are available. What is the shortest possible time to Make Calzones? Let's construct a graph with the nodes and edges first to represent step ordering constraints, and also construct a dictionary to represent time needed for each step. Use the graph and dictionary to calculate the shortest possible time needed for the task. Let's think step by step and then answer the time in double quotes.\\Answer: Here is the adjacency list representation of the step ordering constraints:\\\{'1': ['5'], '2': ['3'], '3': ['4'], '4': ['5'], '5': ['END'], 'END': [], 'START': ['1', '2']\}\\Time for each step can be represented as a dictionary:\\\{'1': '10 min', '2': '10 min', '3': '15 min', '4': '5 min', '5': '25 min'\}\\Since step 1 must precede step 5, step 2 must precede step 3, step 3 must precede step 4, step 4 must precede step 5, we can conclude that we must execute step 2, step 3, step 4, then step 5 sequentially, and since step 1 happens before step 5, it can be done in parallel with step 2, 3, and 4, preceding step 5. Since sequentially executing step 2, 3, 4, and 5 takes 10 + 15 + 5 + 25 = 55 min, while sequentially executing step 1 then step 5 only takes 10 + 25 = 35 min, the shortest possible time to make Calzones is "55 min".\\...\textbf{[TWO MORE EXAMPLES]}...\\\#\#\#\\To create a video game, here are the steps and the times needed for each step.\\Step 1. Learn the basics of programming (180 days)\\Step 2. Learn to use a language that is used in games (60 days)\\Step 3. Learn to use an existing game engine (30 days)\\Step 4. Program the game (90 days)\\Step 5. Test the game (30 days)\\\\\\These ordering constraints need to be obeyed when executing above steps.\\Step 1 must precede step 2.\\Step 1 must precede step 3.\\Step 2 must precede step 4.\\Step 3 must precede step 4.\\Step 4 must precede step 5.\\\\\\Question: Assume that you need to execute all the steps to complete the task and that infinite resources are available. What is the shortest possible time to create a video game? Let's construct a graph with the nodes and edges first to represent step ordering constraints, and also construct a dictionary to represent time needed for each step. Use the graph and dictionary to calculate the shortest possible time needed for the task. Let's think step by step and then answer the time in double quotes.\\Answer:

\colorbox{lightorange}{\textbf{Prototypical task (edge list)}}

The following lists of nodes [0, 1, 2, 3, 4, 5, 6, 7, 8, 9, 10] and edges [[0, 1, 1], [1, 2, 1],[1, 3, 1], [2, 10, 1], ..., [9, 10, 5]] define a directed acyclic graph. Each element in the list of edges is expressed in the form (i,j,w), and specifies that node i connects to node j with weight w. What is the length of the longest path from node 0 to node 10? Think step by step and then reply with the numerical value of the shortest path enclosed by $<$result$>$$<$result$>$ tags.\\Answer:

\subsection{Results after Excluding Invalid Instances}
\label{sec:exclude_invalids}

We report results after excluding invalid instances altogether in all models and settings per experiment if an instance is filtered in any setting in an experiment (e.g. if instance indexed 0 is invalid in zero-shot GPT-4 experiment, we remove it from all the test results for all models and settings). General conclusions remain the same as our main content.

\begin{table*}[h]\centering
    \caption{Model accuracy in different settings on the AsyncHow benchmark. Model performances without our method are in plain background, while those with our method are in blue background. We mark the best performance per model in \textbf{bold}. Following \citet{dror-etal-2018-hitchhikers}, we use McNemar’s tests~\cite{mcnemar1947note} to obtain \emph{p}-values and Holm-Bonferroni method~\cite{holm1979simple} to correct them for each evaluation to test the statistical significance of performance difference between experiment with and without our proposed method. We denote with $\dag$ when the performances with PLaG are significantly better ($p<0.05$) than the best result without.}
    \centering
    \begin{tabular}{c|c|c|c|c|a|a}
        Model & zero-shot & zero-shot + CoT & \emph{k}-shot & \emph{k}-shot + CoT& PLaG (explicit graph) & PLaG (BaG)\\\hline
        GPT-4 & 0.128 & 0.128 & 0.108 & 0.657 & 0.728$^\dag$ &\textbf{0.771}$^\dag$\\
        GPT-3.5 & 0.191 & 0.217& 0.241&0.224&0.284$^\dag$&\textbf{0.348}$^\dag$\\
        Command & 0.079 & 0.015& 0.051& 0.078&\textbf{0.098}&0.052\\
        LLaMA-2-70B-chat & 0.039& 0.036&0.053 & 0.074&\textbf{0.101}$^\dag$&0.069\\
        Mistral-7B-Instruct&0.074&0.070&0.099&0.142&\textbf{0.155}&0.144\\
    \end{tabular}
    \label{tab:exclude_invalids-app}
\end{table*}

\begin{figure}[h]
     \centering
    \includegraphics[scale=0.7]{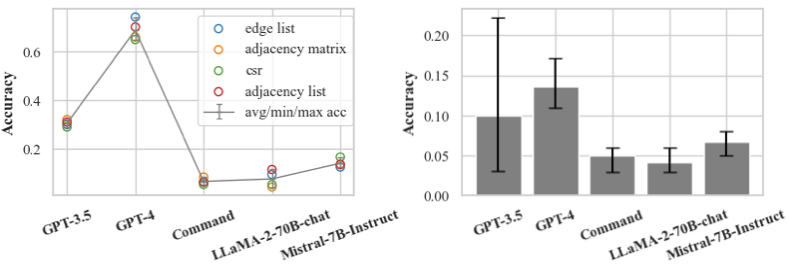}
    \caption{\small Left plot refers to average model performance accuracy with k shot + CoT in different graphs. Dots refer to model performance with different graph types. Grey lines refer to average accuracy with different graph types. The right plot refers to average zero-shot accuracy in different text prompts. Error bars in both plots refer to worst/best performance}
    \label{fig:vary-prompt-graph-app}
\end{figure}

\begin{figure}[h]
    \centering
    \includegraphics[scale=0.7]{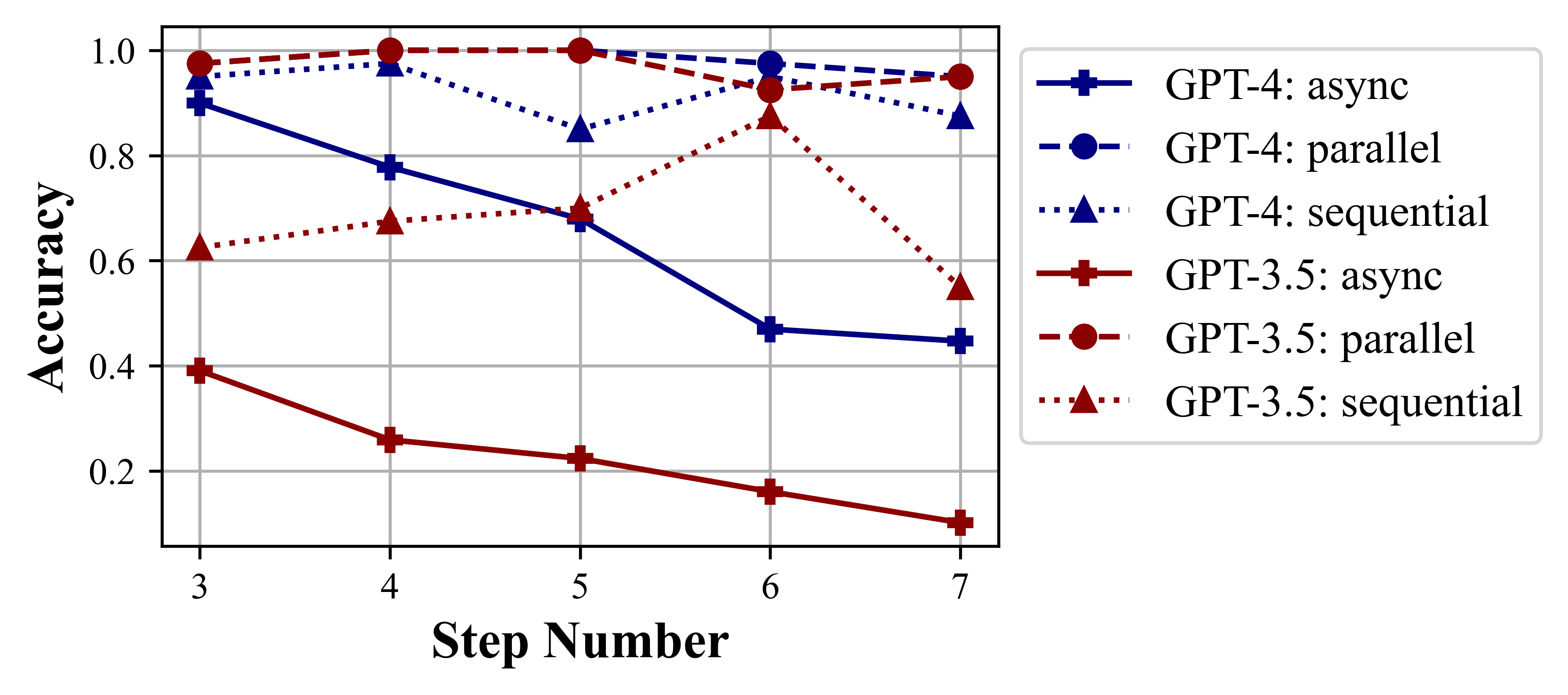}
    \caption{Comparing parallel/sequential plan execution accuracy with asynchronous plans. All experiments are done in the setting of \emph{k}-shot + CoT. Blue and orange lines refer to GPT-4 and GPT-3.5 results respectively.}
    \setlength{\belowcaptionskip}{-100pt}
    \label{fig:baseline-app}
\end{figure}

\begin{figure}[!h]
    \centering
    \includegraphics[scale=0.8]{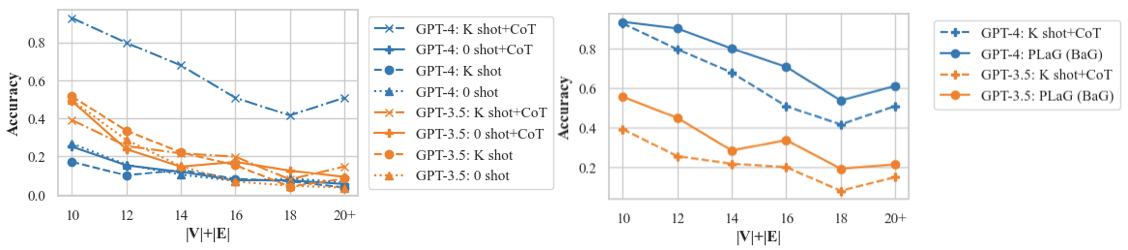}
    \caption{Model accuracy concerning task complexity. The left figure plots model performance without our method (PLaG), and the right plot displays the models' best performance with/without our method.}
    \label{fig:acc-ve-appendix}
\end{figure}

\newpage
\subsection{Further Comparison with Chain-of-Thought Self-consistency and Tree of Thought}
\label{sec:sc_tot}
We further added experiments on Chain-of-Thought Self-Consistency (CoT-SC) \cite{wang2022self} and ToT \cite{yao2024tree} (both for k shot, k=3 as in our main paper). We report results for k-shot CoT-SC and ToT, in comparison to PLaG below (best results are in bold). We see that our method outperforms both methods while inducing much less cost (see Appendix~\ref{sec:latency}).

\begin{table}[h!]
    \centering
    \caption{Comparing \emph{k}-shot CoT-SC/ToT with PLaG on GPT-3.5 and GPT-4. The best results per model are in bold. Our methods are always superior.}
    \begin{tabular}{c|c|c|c|c}
         &\emph{k}-shot CoT-SC&\emph{k}-shot ToT&PLaG (explicit graph)&PLaG (BaG)\\\hline
         GPT-3.5&0.240&0.263&0.290&\textbf{0.355}\\
         GPT-4&0.625&0.624&0.730&\textbf{0.777}
    \end{tabular}
    \label{tab:sc_tot}
\end{table}

\subsection{Latency Analysis}
\label{sec:latency}
We analyze the cost-performance trade-off for PLaG here. PLaG introduces significantly longer inference sequences than the zero-shot setting, but we note that its cost is reasonable compared to other advanced prompting methods. We provide a statistical comparison of the average input/output token count per task for GPT-4 below. 

We note that the output token lengths of PLaG (explicit graph) are comparable to that of \emph{k}-shot CoT, while \emph{k}-shot CoT-SC and \emph{k}-shot ToT are much more expensive and underperform both PLaG methods (see Appendix~\ref{sec:sc_tot}).

\begin{table}[h!]
    \centering
    \caption{Latency analysis: comparing PLaG with other prompting methods.}
    \begin{tabular}{c|c|c|c|c|c|c}
        &zero shot&\emph{k}-shot CoT&PLaG (explicit graph)&PLaG (BaG)&\emph{k}-shot CoT-SC&\emph{k}-shot ToT \\\hline
         tokens (input/output)&207/5&1289/135&1698/138&1775/242&1289/407&5212/335
    \end{tabular}
    \label{tab:latency}
\end{table}

The increased input and output length caused by PLaG may raise concerns about the scaling potential of the technique (i.e., whether it is still applicable when there are hundreds of nodes or more in a graph). First, we consider this to be a problem mainly for LLM context window length, which is universal to all NLP problems in general. In summarization tasks, for example, an LLM can't summarize a book whose length goes beyond the LLM's context length, but this doesn't invalidate summarization as a task. Second, the length of graph representation also depends on the graph format: if a graph is provided as its dependency list, its size will grow linearly with the number of nodes and edges, as opposed to the adjacency matrix, which scales quadratically.

Second, our dataset, which is generated from real-life tasks without specific pruning for complex ones, shows very sparse data points for complexity which goes beyond 20 (Figure~\ref{fig:data-distribution}). This observation motivates us to consider additional graph prompting as a valid technique to improve LLM performance in a wide range of tasks.

\subsection{Comparing PLaG Performance on Model-generated and Human-annotated Data}
\label{sec:plag_noise}
Here, we compare GPT-3.5/4 performance on model-generated and human-annotated data to show that the performance gap between the explicit graph and BaG used in PLaG is not caused by noise in sampling. We perform additional experiments to compare BaG and explicit graph on task complexity 14, which has $>100$ data points for both the synthetic and human-annotated parts of our dataset. We find that the BaG framework performs consistently better than the explicit graph as shown below, which means the superior performance of BaG should not be attributed to noise sampling.

\begin{table}[h!]
    \centering
    \caption{Compare explicit graph (left) and BaG (right) as PLaG methods in human-annotated data and model-generated data. The best results are in bold.}
    \begin{tabular}{c|c|c}
    & Human-annotated data	& Generated data\\\hline
GPT-3.5 (explicit graph/BaG)&0.239/\textbf{0.279}	&0.315/\textbf{0.389}\\
GPT-4 (explicit graph/BaG)&0.614/\textbf{0.701}&0.778/\textbf{0.796}
    \end{tabular}
    \label{tab:plag_noise}
\end{table}

\subsection{Model Performance with Economic Linguistic Expressions}
\label{sec:economic_exp}
We report results comparing LLMs' performance between direct and economic expressions in Figure~\ref{tab:economic_exp}. Generally, models' performance downgrades when using economic expressions except LLaMA-2-70B-chat.
\begin{table}[h!]
    \caption{Compare model performance with unambiguous direct expression with economic expressions.}
    \centering
    \begin{tabular}{c|c|c}
        Model & Best direct expression performance & Performance difference after using economic expressions \\\hline
        GPT-3.5 &0.222 &-0.012\\
        GPT-4 & 0.171&-0.021\\
        Command & 0.06&-0.01\\
        LLaMA-2-70B-chat & 0.06&+0.06\\
        Mistral-7B-Instruct & 0.08&-0.01\\
    \end{tabular}
    \label{tab:economic_exp}
\end{table}

\newpage
\subsection{The \emph{Prototypical} Distribution as a Proxy of the Naturalistic Benchmark}
\label{sec:distr-ood}
\begin{figure}[h!]
    \centering
    \includegraphics[width=0.8\linewidth]{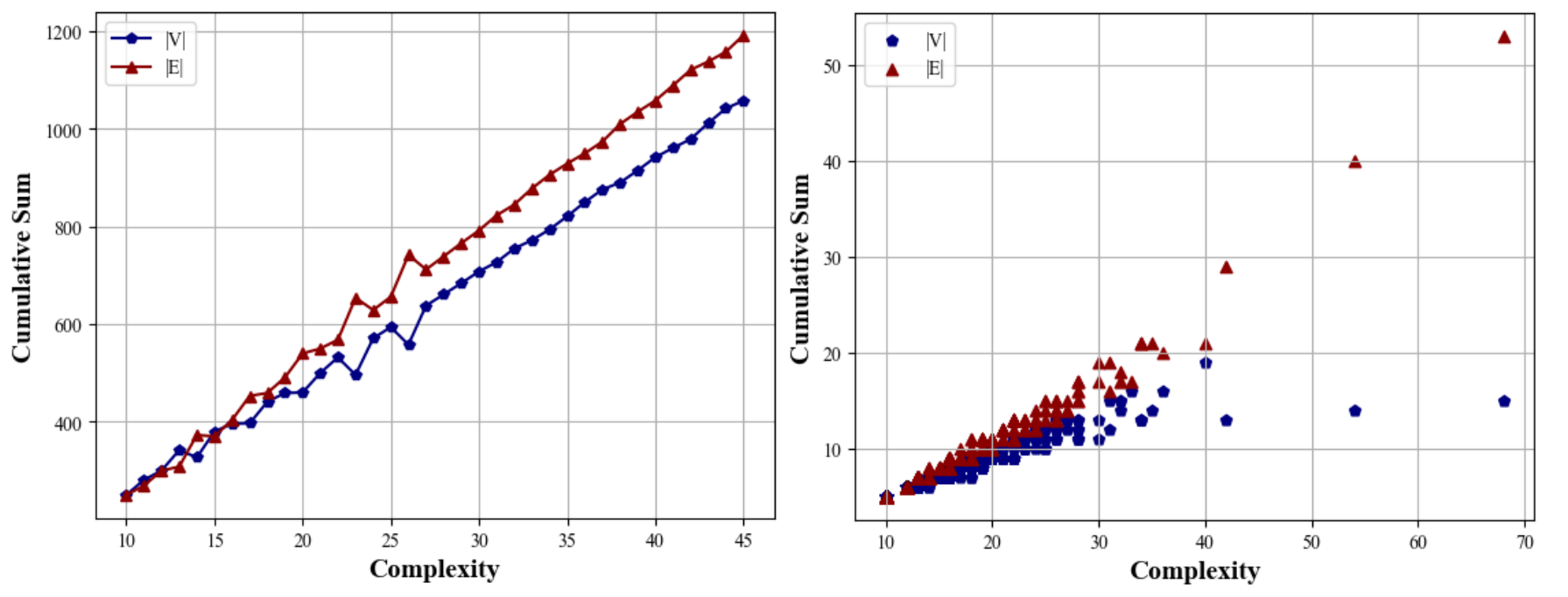}
    \caption{\small Comparison of the cumulative number of vertices $|V|$ and edges $|E|$ per-complexity task for the prototypical (left) vs. naturalistic datasets.}
    \label{fig:proto-vs-nat-complexity}
\end{figure}

\begin{figure}[h!]
    \centering
    \includegraphics[width=0.5\linewidth]{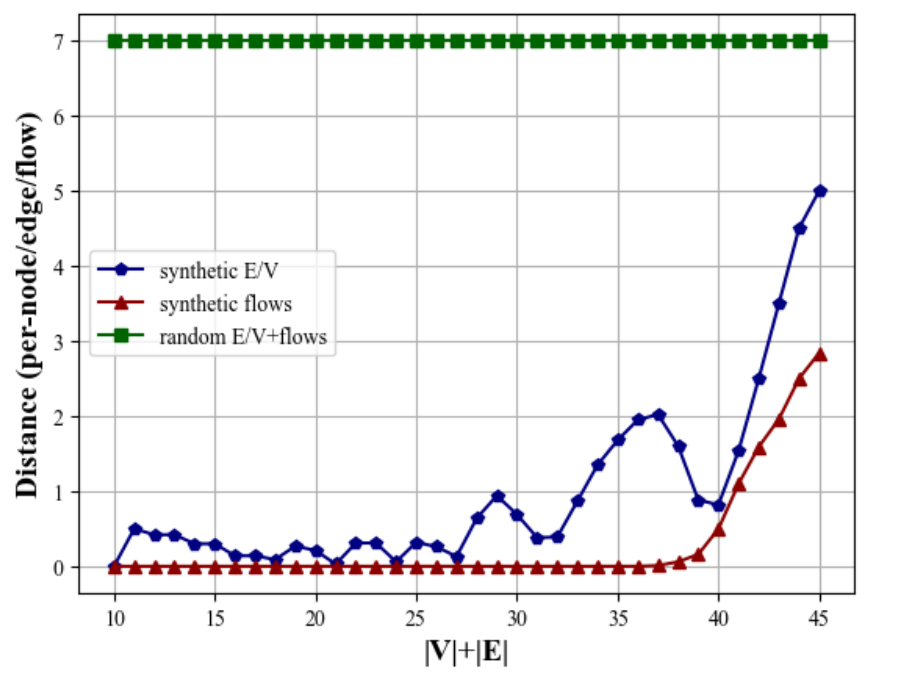}
    \caption{\small Comparison of the graph distribution of prototypical graphs and that of random graphs. In \textcolor{ceruleanblue}{blue} and \textcolor{purple}{red}, a comparison between the number of nodes/edges and flows of prototypical and naturalistic graphs. In \textcolor{teal}{green}, comparison of nodes/edges and flows of prototypical and random graphs (baseline). For larger complexities, prototypical graphs become similar to DAG in terms of the number of edges, nodes and flows (i.e., tasks that can be executed simultaneously).}
    \label{fig:distr-ood}
\end{figure}

In this subsection, we show that $|V|$ and $|E|$ equally contribute to the complexity of a planning task, with no clear dominance of one over the other Figure~\ref{fig:proto-vs-nat-complexity} and the similarity of prototypical and naturalistic graphs Figure~\ref{fig:distr-ood}.

\subsection{Analysing Time Units Differences in Complexity Levels $20+$ and $18$}
\label{sec:time_unit}
We define a list of time units ['sec', 'min', 'h', 'day', 'week', 'month', 'year'] and define time unit distance per instance as the difference between the unit with the highest and lowest index. For instance, a script with steps timed as 5 sec, and 10 min respectively is considered as having a distance of 1, while a script with steps timed as 15 h and 50 h has a distance of 0. 

We find that the average time distance over all scripts at $20+$ (0.339) is considerably lower than the average distance at complexity $18$ (0.801), which partially explains why the accuracy at $20+$ has a jump.

\subsection{Experiment Details and Hyperparameters}
All experiments are performed from December 2023 to May 2024.

For data generation, we use Azure OpenAI API and set temperature=1 for both GPT-35-turbo and GPT-4. In dependency validation, we sample prompts by seed 0, 1, and 2.

During the experiment (i.e. inference stage), we use Azure OpenAI API and set temperature=0 for GPT models to enable as much reproducibility as possible. We use Cohere API to query the Command model and also set temperature = 0. As GPT models and Command filter contents, we query API 3 times to see if the corresponding model is willing to answer the prompt.

We use Huggingface Inference API to query LLaMA-70B-Chat and set do sample=False, max new tokens=4096, and seed=0. We use 2 V100 GPUs and 1 A100 GPU for Mistral-7B-instruct inference, with do\_sample=False, temperature=0, max new tokens=4096 and torch manual seed=2024. 

\subsection{Dataset Information}
We use ProScript and WikiHow as our base dataset in data generation. We follow the licensing guide of ACL and determine ProScript to be under CC BY 4.0. WikiHow dataset we use is under MIT License. We follow the licenses used by the existing datasets for our dataset.

\subsection{Human Validation}
We conduct human validation of WikiHow on a voluntary basis with four experts. Consent was obtained via discussion with them. We do not provide personally identifiable information in the dataset.

\subsection{Statement of Contribution}

FL wrote the paper, developed the initial idea, developed part of formalism, generated the AsyncHow dataset, ran all experiments, and conducted all analyses unless specified below. EMY helped in polishing ideas, rephrasing prompts, and editing the paper. ELM developed part of formalism, generated a synthetic dataset, ran synthetic experiments, and advised and edited the paper. AGC obtained funding, advised, and edited the paper. VH and JBP advised and edited the paper.

\end{document}